\documentclass[twoside]{article}
\usepackage[accepted]{aistats2021}
\usepackage[utf8]{inputenc}
\usepackage[T1]{fontenc}
\usepackage{hyperref} 
\usepackage{url}      
\usepackage{booktabs} 
\usepackage{amsfonts} 
\usepackage{nicefrac} 
\usepackage{microtype}
\usepackage{amsmath}
\usepackage{amssymb}
\usepackage{amsthm}
\usepackage{mathrsfs}
\usepackage{mathtools}
\usepackage{dsfont}
\usepackage{enumerate}
\usepackage[,slantedGreek]{mathptmx}
\usepackage{subfig}
\usepackage{subfig}
\usepackage{float}
\usepackage{natbib}
\bibpunct{(}{)}{;}{a}{,}{,}

\hypersetup{
colorlinks = true,
urlcolor = blue,
linkcolor = blue,
citecolor = blue,
}

\newtheorem{theorem}{Theorem}
\newtheorem{assumption}{Assumption}
\newtheorem{definition}{Definition}

\newtheorem{lemma}{Lemma}

\newtheorem{proposition}{Proposition}
\newtheorem{corollary}{Corollary}

\begin{document}

\twocolumn[
\aistatstitle{Approximating Lipschitz continuous functions with 
\\ GroupSort neural networks}
\aistatsauthor{  U. Tanielian \\
  Criteo, Sorbonne Universit\'e\\
  \small{{\tt u.tanielian@criteo.com}} \\\And
  M. Sangnier \\
  Sorbonne Universit\'e\\
  \small{{\tt maxime.sangnier@upmc.fr}} \\\And
   G. Biau \\
  Sorbonne Universit\'e\\
  \small{{\tt gerard.biau@upmc.fr}} }
\aistatsaddress{Paris, France \And  Paris, France \And Paris, France}]

\begin{abstract}
    Recent advances in adversarial attacks and Wasserstein GANs have advocated for use of neural networks with restricted Lipschitz constants. Motivated by these observations, we study the recently introduced GroupSort neural networks, with constraints on the weights, and make a theoretical step towards a better understanding of their expressive power. We show in particular how these networks can represent any Lipschitz continuous piecewise linear functions. We also prove that they are well-suited for approximating Lipschitz continuous functions and exhibit upper bounds on both the depth and size. To conclude, the efficiency of GroupSort networks compared with more standard ReLU networks is illustrated in a set of synthetic experiments.
\end{abstract}

\section{Introduction}\label{section:introduction}
In the past few years, developments in deep learning have highlighted the benefits of operating neural networks with restricted Lipschitz constants. An important illustration is provided by robust machine learning, where networks with large Lipschitz constants are prone to be more sensitive to adversarial attacks, in the sense that small perturbations of the inputs can lead to significant misclassification errors \citep[e.g.,][]{goodefellow2015explaining}. In order to circumvent these limitations, \citet{gao2017wasserstein}, \citet{esfahani2018data}, and \citet{blanchet2019robust} studied a new regularization scheme based on penalizing the gradients of the networks. Constrained neural networks also play a key role in the different but not less important domain of Wasserstein GANs \citep{arjovsky2017wasserstein}, which take advantage of the dual form of the $1$-Wasserstein distance expressed as a supremum over the set of $1$-Lipschitz functions \citep{villani2008optimal}. This formulation has been shown to bring training stability and is empirically efficient \citep{gulrajani2017improved}. In this context, many different ways have been explored to restrict the Lipschitz constants of the discriminator. One possibility is to clip their weights, as advocated by \citet{arjovsky2017wasserstein}. Other solutions involve enforcing a gradient penalty \citep{gulrajani2017improved} or penalizing norms of the matrices of the weights \citep{spectral_normGANs}. 

However, all of these operations are delicate and may significantly affect the expressive power of the neural networks. For example, \citet[][]{huster2018limitations} show that ReLU neural networks with constraints on the weights cannot represent even the simplest functions, such as the absolute value. In fact, little is known regarding the expressive power of such restricted networks, since most studies interested in the expressiveness of neural networks \citep[e.g.,][]{hornik1989multilayer, cybenko1989approximation, raghu2017expressive} do not take into account eventual constraints on their architectures. As far as we know, the most recent attempt to tackle this issue is by \citet{Anil2018SortingOL}. These authors exhibit a family of neural networks, with constraints on the weights, which is dense in the set of Lipschitz continuous functions on a compact set. To show this result, \citet{Anil2018SortingOL} make critical use of GroupSort activations.

Motivated by the above, our objective in the present article is to make a step towards a better mathematical understanding of the approximation properties of Lipschitz feedforward neural networks using GroupSort activations. Our contributions are threefold:
\begin{enumerate}[$(i)$]
    \item We show that GroupSort neural networks, with constraints on the weights, can represent any Lipschitz continuous piecewise linear function and exhibit upper bounds on both their depth and size. We make a connection with the literature on the depth and size of ReLU networks \citep[in particular][]{arora2018understanding, he2018relu}. 
    \item Building on the work of \citet{Anil2018SortingOL}, we offer upper bounds on the depth and size of GroupSort neural networks that approximate $1$-Lipschitz continuous functions on compact sets. We also show that increasing the grouping size may significantly improve the expressivity of GroupSort networks. 
    \item We empirically compare the performances of GroupSort and ReLU networks in the context of function regression estimation and Wasserstein distance approximation. 
\end{enumerate}
The mathematical framework together with the necessary notation is provided in Section \ref{section:background}. Section \ref{sec:grouping_size_2} is devoted to the problem of representing Lipschitz continuous functions with GroupSort networks of grouping size $2$. The extension to any arbitrary grouping size is discussed in Section \ref{sec:grouping_size_k} 
and numerical illustrations are given in Section \ref{section:experiments}. For the sake of clarity, all proofs are gathered in the Appendix.

\section{Mathematical context}\label{section:background}
We introduce in this section the mathematical context of the article and describe more specifically the GroupSort neural networks, which, as we will see, play a key role in representing and approximating Lipschitz continuous functions.

Throughout the paper, the ambient space $\mathds R^d$ is assumed to be equipped with the Euclidean norm $\|\cdot\|$. For $E$ a subset of $\mathds R^d$, we denote by $\text{Lip}_1(E)$ the set of $1$-Lipschitz real-valued functions on $E$, i.e.,
\begin{equation*}
    \text{Lip}_1(E) = \big\{ f :E \to \mathds R: |f(x)-f(y)|\leqslant {\|x-y\|}, \ (x,y) \in E^2\big\}
\end{equation*}
Let $k \geqslant 2$ be an integer. We let $\mathscr{D}_k = \{D_{k,\alpha}: \alpha \in \Lambda \}$ be the class of functions from $\mathds R^d$ to $\mathds R$ parameterized by feedforward neural networks of the form
\begin{align}\label{eq:def_discriminators}
    D_{k, \alpha}(x) &= \underset{1 \times v_{q-1}}{V_{q}} \sigma_k (\underset{v_{q-1} \times v_{q-2}}{V_{q-1}} \cdots \sigma_k (\underset{v_2 \times v_1}{V_2} \sigma_k (\underset{v_1 \times D}{V_1} x + \underset{v_1\times 1}{c_1}) \nonumber \\ 
    &+ \underset{v_2 \times 1}{c_2}) + \underset{v_{q-1}\times 1}{c_{q-1}}) + \underset{1 \times 1}{c_q},
\end{align}
where $q \geqslant 2$ and the characters below the matrices indicate their dimensions ($\mbox{lines} \times \mbox{columns}$). For $q=1$, we simply let $D_{k,\alpha}(x) = V_1x+c_1$ be a simple linear regression in $\mathds{R}$ without hidden layers. Thus, a network in $\mathscr{D}_k$ has $(q-1)$ hidden layers, and hidden layers from depth $1$ to $(q-1)$ are assumed to be of respective widths $v_i$, $i=1, \hdots, q-1$, \textit{divisible by $k$}. Such a network is said to be of depth $q$ and of size $\nu_1 + \cdots +\nu_{q-1}$. The matrices $V_i$ are the matrices of weights between layer $i$ and layer $(i+1)$ and the $c_i$'s are the corresponding offset vectors (in column format). So, altogether, the vectors $\alpha=(V_1, \hdots, V_q, c_1, \hdots, c_q)$ represent the parameter space $\Lambda$ of the functions in $\mathscr{D}_k$. 
With respect to the activation functions $\sigma_k$, we propose to use the GroupSort activation, which separates the pre-activations into groups and then sorts each group into ascending order. 

The GroupSort function splits the input into $n$ different groups of $k$ elements each: $G_1 = \{x_1, \hdots, x_k \} , \hdots, G_n = \{x_{nk-(k-1)}, \hdots, x_{nk}\}$, and then orders each group by decreasing order. Thus, the GroupSort function with a grouping size $k\geqslant2$ is applied on a given vector $(x_1, \hdots , x_{kn})$ as follows:
\begin{align*}
    {\sigma_k}(x_1, \hdots, x_k, \hdots, x_{nk-(k-1)}, \hdots, x_{nk}) = \\ \big(x_{(k)}^{G_1}, \hdots, x_{(1)}^{G_1}, \hdots, x_{(k)}^{G_n}, \hdots, x_{(1)}^{G_n} \big),
\end{align*}
where $(x_{(k)}^{G_i}, \hdots, x_{(1)}^{G_i})$ corresponds to the decreasing ordering in the group $G_i$.
\begin{figure}
    \centering
    {
        \includegraphics[width=0.95\linewidth]{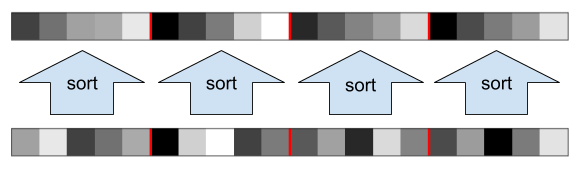}
    }\caption{GroupSort activation with a grouping size 5. Source: \citet{Anil2018SortingOL}.}
    \label{fig:groupsort_k5}
\end{figure}

This activation is applied on groups of $k$ components, which makes sense in \eqref{eq:def_discriminators} since the widths of the hidden layers are assumed to be divisible by $k$. GroupSort has been introduced in \citet{Anil2018SortingOL} as a $1$-Lipschitz activation function that preserves the gradient norm of the input. An example with a grouping size $k=5$ is given in Figure \ref{fig:groupsort_k5}. With a slight abuse of vocabulary, we call a neural network of the form \eqref{eq:def_discriminators} a GroupSort neural network. We note that the GroupSort activation can recover the standard rectifier function. For example, $\sigma_2(x, 0) = (\text{ReLU}(x), -\text{ReLU}(-x))$, but the converse is not true. 

Throughout the manuscript, the notation $\|\cdot\|$ (respectively, $\|\cdot\|_{\infty}$) means the Euclidean (respectively, the supremum) norm on $\mathds R^p$, with no reference to $p$ as the context is clear. For $W=(w_{i,j})$ a matrix of size $p_1 \times p_2$, we let $\|W\|_2 = \sup_{\|x\|=1} \|Wx\|$ be the $2$-norm of $W$. Similarly, the $\infty$-norm of $W$ is $\|W\|_\infty = \sup_{\|x\|_{\infty}=1} \|Wx\|_{\infty}=\max_{i=1, \hdots, p_1} \sum_{j=1}^{p_2} |w_{i,j}|$. We will also use the $(2,\infty)$-norm of $W$, i.e., $\|W\|_{2, \infty} = \sup_{\|x\|=1} \|Wx\|_\infty$. The following assumption plays a central role in our approach: 
\begin{assumption}\label{ass:compactness}
    For all $\alpha=(V_1, \hdots, V_q, c_1, \hdots, c_q) \in \Lambda$,
    \begin{align*}
        \|V_1\|_{2, \infty} \leqslant 1 &, \ \max(\|V_2\|_\infty, \hdots, \|V_q\|_\infty) \leqslant 1, \ \\
        &\emph{and} \ \max(\|c_i\|_\infty: i = 1,\hdots,q) \leqslant K_2,
    \end{align*}
    where $K_2 \geqslant 0$ is a constant.
\end{assumption}
This type of compactness requirement has already been suggested in the statistical and machine learning community \citep[e.g.,][]{arjovsky2017wasserstein, Anil2018SortingOL, biau2018some}. In the setting of this article, its usefulness is captured in the following simple but essential lemma:
\begin{lemma}\label{lem:uniformly_lipschitz_neural_nets}
    Assume that Assumption \ref{ass:compactness} is satisfied. Then, for any $k \geqslant 2$, $\mathscr{D}_k \subseteq \emph{Lip}_1(\mathds R^d)$.
\end{lemma} 
Combining Lemma \ref{lem:uniformly_lipschitz_neural_nets} with Arzel\`a-Ascoli theorem, it is easy to see that, under Assumption \ref{ass:compactness}, the class $\mathscr{D}_k$ restricted to any compact  $K \subseteq \mathds R^d$ is compact in the set of continuous functions on $K$ with respect to the uniform norm. From this point of view, Assumption \ref{ass:compactness} is therefore somewhat restrictive. On the other hand, it is essential in order to guarantee that all neural networks in $\mathscr{D}_k$ are indeed 1-Lipschitz. Practically speaking, various approaches have been explored in the literature to enforce this $1$-Lipschitz constraint. \citet{gulrajani2017improved}, \citet{kodali2017convergence}, \citet{wei2018improving}, and \citet{zhou2019lipschitz} proposed a gradient penalty term, \citet{spectral_normGANs} applied spectral normalization, while \citet{Anil2018SortingOL} have shown the empirical efficiency of the orthonormalization of \citet{bjorck1971iterative}.

Importantly, \citet[][Theorem 3]{Anil2018SortingOL} states that, under Assumption \ref{ass:compactness}, GroupSort neural networks are universal Lipschitz approximators on compact sets. More precisely, for any Lipschitz continuous function $f$ defined on a compact, one can find a neural network of the form \eqref{eq:def_discriminators} verifying Assumption \ref{ass:compactness} and arbitrarily close to $f$ with respect to the uniform norm. Our objective in the present article is to explore the properties of these networks. We start in the next section by examining the case of piecewise linear functions.

\section{Learning functions with a grouping size 2} \label{sec:grouping_size_2}
For this section, we only consider GroupSort neural networks with a grouping size  2 and aim at studying their expressivity. The capacity of GroupSort networks to approximate continuous functions is studied via the representation of piecewise linear functions. For feedforward ReLU networks, their ability to represent such functions has been largely studied. In particular, \citet[][Theorem 2.1]{arora2018understanding} reveals that any piecewise linear function from $\mathds{R}^d \to \mathds{R}$ can be represented by a ReLU network of depth at most $\lceil\log_2(d+1)\rceil$ (the symbol $\lceil \cdot \rceil$ stands for the ceiling function), whereas \citet{he2018relu} specify an upper bound on their size. In the present section, we extend these results and first tackle the problem of representing piecewise linear functions with constrained GroupSort networks. Then we move to the non-linear case. 
\subsection{Representation of piecewise linear functions}
\begin{figure}
    \centering
    {
        \includegraphics[width=0.80\linewidth]{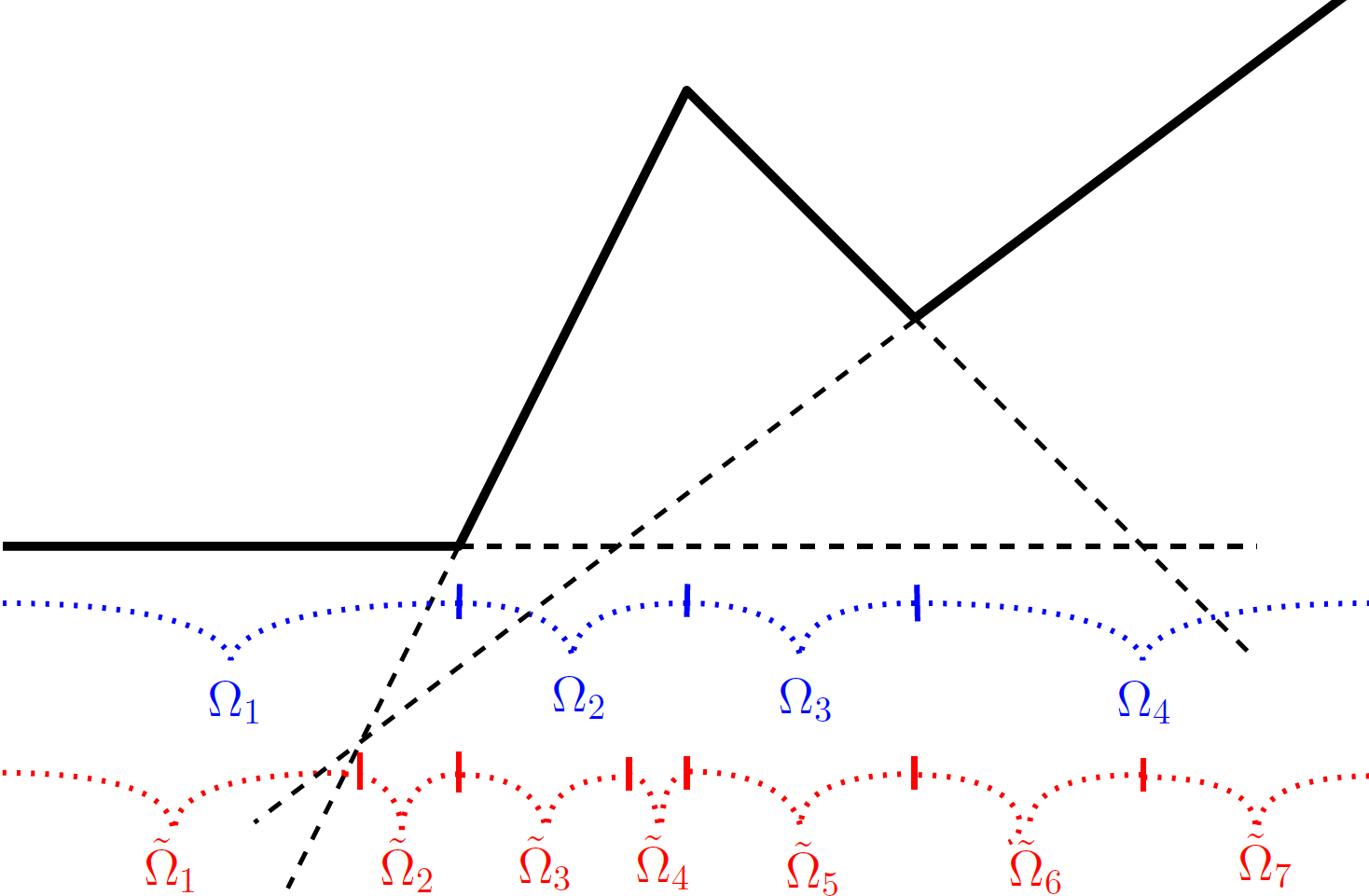}
    }\caption{A $4$-piecewise linear function on the real line and the associated partitions $\Omega=\{\Omega_1, \hdots, \Omega_4\}$ and $\Tilde{\Omega}=\{\Tilde{\Omega}_1, \hdots, \Tilde{\Omega}_{7}\}$. The partition $\Tilde{\Omega}$ is finer than $\Omega$.}
    \label{fig:omega_and_omegatilde}
\end{figure}

Let us start gently by fixing the vocabulary.
\begin{definition}
    A continuous function $f:\mathds{R}^d \to \mathds{R}$ is said to be (continuous) $m_f$-piecewise linear ($m_f \geqslant 2$) if there exist a partition $\Omega =\{\Omega_1, \hdots, \Omega_{m_f}\}$ of $\mathds{R}^d$ into polytopes and a collection $\ell_1, \hdots, \ell_{m_f}$ of affine functions such that, for all $x \in \Omega_i$, $i=1, \hdots, m_f$, $f(x)=\ell_i(x)$.
\end{definition}
At this stage no further assumption is made on the sets $\Omega_1, \hdots, \Omega_{m_f}$, which are just assumed to be polytopes in $\mathds{R}^d$. An example of piecewise linear function on the real line with $m_f=4$ is depicted in Figure \ref{fig:omega_and_omegatilde}. 
As this figure suggests, the ambient space $\mathds{R}^d$ can be further covered by a second partition $\Tilde{\Omega}=\{\Tilde{\Omega}_1, \hdots, \Tilde{\Omega}_{M_f}\}$ of $M_f$ polytopes ($M_f\geqslant 1$), in such a way that the sign of the differences $\ell_i - \ell_j$, $(i,j) \in \{1,\hdots,m_f\}^2$, does not change on the subsets $\Tilde{\Omega}_1, \hdots, \Tilde{\Omega}_{M_f}$. It is easy to see that the partition $\Tilde{\Omega}$ is finer than $\Omega$ since, for each $i \in \{1,\hdots, M_f\}$ there exists $j \in \{1, \hdots, m_f\}$ such that $\Tilde{\Omega_i} \subseteq \Omega_j$. This implies in particular that $M_f \geqslant m_f$. 

The usefulness of the partition $\Tilde{\Omega}$ is demonstrated by \citet[][Theorem 5.1]{he2018relu}, which states that any $m_f$-piecewise linear function $f$ can be written as
\begin{equation}\label{eq:piecewise_linear_characterization}
    f = \underset{1 \leqslant k \leqslant M_f}{\max} \ \underset{i \in S_k}{\min} \ \ell_i,
\end{equation}
where each $S_k$ is a non-empty subset of $\{1, \hdots, m_f\}$. This characterization of the function $f$ is interesting, since it shows that any $m_f$-piecewise linear function can be computed using only a finite number of $\max$ and $\min$ operations. As identity \eqref{eq:piecewise_linear_characterization} is essential for our approach, this justifies spending some time examining it.
\begin{lemma}\label{lem:number_of_ordered_subdomains}
    Let $f: \mathds{R}^d \to \mathds{R}$ be an $m_f$-piecewise linear function. Then $m_f \leqslant M_f \leqslant \min(2^{m_f^2/2}, (m_f/\sqrt{2})^{2d})$.
\end{lemma}
Lemma \ref{lem:number_of_ordered_subdomains} is an improvement of \citet[][Lemma 5.1]{he2018relu}, which shows that $M_f \leqslant m_f!$. Our proof method exploits the inequality $M_f \leqslant C_{m_f(m_f-1)/2, d}$, where $C_{n,d}$ denotes the number of arrangements of $n$ hyperplanes in a space of dimension $d$ \citep[][Chapter 5]{devroye2013probabilistic}. Another application of \eqref{eq:piecewise_linear_characterization} is encapsulated in Proposition \ref{lem:auxiliary_lemma_max_function} below, which will be useful for later analysis, in combining maxima and minima in neural networks of the form \eqref{eq:def_discriminators}.
\begin{proposition}\label{lem:auxiliary_lemma_max_function}
    Let $f_1, \hdots, f_m: \mathds{R}^d \to \mathds{R}$ be a collection of functions ($m \geqslant 2$), each represented by a neural network of the form \eqref{eq:def_discriminators}, with common depth $q$ and sizes $s_i$,  $i=1, \hdots, m$. 
    
    In the specific case where $m=2^n$ for some $n \geqslant 1$, there exist neural networks of the form \eqref{eq:def_discriminators} (with grouping size $2$) with depth $q+\log_2(m)$ and size at most $s_1 + \cdots + s_{m} + m -1$ that represent the functions $f= \max(f_1, \hdots, f_{m})$ and $g= \min(f_1, \hdots, f_{m})$. 
    
    If $m$ is arbitrary, then there exist neural networks of the form \eqref{eq:def_discriminators} with depth $q + \lceil\log_2(m) \rceil$ and size at most $s_1 + \cdots + s_{m} + 2m -1$ that represent the functions $f$ and $g$. 
    
\end{proposition}
Interestingly, \citet[][Lemma D.3]{arora2018understanding}, which is the analog of Proposition \ref{lem:auxiliary_lemma_max_function} asserts that the size with ReLU activations is at most $s_1 + \cdots + s_m + 8m -4$. For the specific computation of maxima/minima of functions, it should be stressed that GroupSort activations slightly reduces the size of the networks. By combining Lemma \ref{lem:number_of_ordered_subdomains}, Proposition \ref{lem:auxiliary_lemma_max_function}, and identity \eqref{eq:piecewise_linear_characterization}, we are led to the following theorem, which reveals the ability of GroupSort networks for representing $1$-Lipschitz piecewise linear functions.
\begin{theorem}\label{th:disc_can_represent_piecewise_linear}
    Let $f \in \emph{Lip}_1(\mathds R^d)$ that is also $m_f$-piecewise linear. Then there exists a neural network of the form \eqref{eq:def_discriminators} verifying Assumption \ref{ass:compactness} that represents $f$. Besides, its depth is $\lceil \log_2(M_f) \rceil + \lceil \log_2(m_f) \rceil +1$ and its size is at most $3m_f M_f+M_f-1$. 
\end{theorem}
This result should be compared with state-of-the-art results known for ReLU neural networks. In particular, \citet[][Theorem 2.1]{arora2018understanding} reveals that any $m_f$-piecewise linear function $f$ can be represented by a ReLU network with depth at most $\lceil\log_2(d+1) \rceil$. The upper bound of Theorem \ref{th:disc_can_represent_piecewise_linear} can be larger since it involves both $M_f$ and $m_f$. On the other hand, the upper bound $O(m_f M_f)$ on the size significantly improves on \citet[][Theorem 5.2]{he2018relu}, which is at least $O(d2^{m_f M_f})$. This improvement in terms of size can be roughly explained by the depth/size trade-off results known in deep learning theory. As a matter of fact, many theoretical research papers have underlined the benefits of depth relatively to width for parameterizing complex functions \citep[as, for example, in][]{telgarsky2015representation, telgarsky2016benefits}. For a fixed number of neurons, when comparing two neural networks, the deepest is the most expressive one \citep{lu2017expressive}. 

It turns out that Theorem \ref{th:disc_can_represent_piecewise_linear} can be significantly refined when the partition $\Omega$ satisfies some geometrical properties. Our next proposition examines the case where the sets $\Omega_1, \hdots, \Omega_{m_f}$ are convex.
\begin{corollary}\label{cor:piecewise_linear_functions_on_convex_sets}
Let $f \in \emph{Lip}_1(\mathds R^d)$ that is also $m_f$-piecewise linear with convex subdomains $\Omega_1, \hdots, \Omega_{m_f}$. Then there exists a neural network of the form \eqref{eq:def_discriminators} verifying Assumption \ref{ass:compactness}  that represents $f$. Besides, its depth is $2\lceil \log_2(m_f) \rceil+1$ and its size is at most $3m_f^2+m_f-1$. 
\end{corollary}
Corollary \ref{cor:piecewise_linear_functions_on_convex_sets} offers a significant improvement over Theorem \ref{th:disc_can_represent_piecewise_linear}, since in general $M_f \gg m_f$. We note in passing that the result of this proposition is dimension-free. 

\subsection{GroupSort neural networks on the real line}
Piecewise linear functions defined on $\mathds{R}$ deserve a special treatment, since in this case, any connected subset is convex. 
\begin{proposition}\label{cor:func_r_to_r}
    Let $f \in \emph{Lip}_1(\mathds R)$ that is also $m_f$-piecewise linear. Then there exists a neural network of the form \eqref{eq:def_discriminators} verifying Assumption \ref{ass:compactness}  that represents $f$. Besides, its depth is $2 \lceil \log_2(m_f) \rceil+1$ and its size is at most $3m_f^2+m_f-1$. 
    
    In the specific case where $f$ is convex (or concave), then there exists a neural network of the form \eqref{eq:def_discriminators} verifying Assumption \ref{ass:compactness}  that represents $f$. Its depth is $\lceil \log_2(m_f) \rceil+1$ and its size is at most $3m_f-1$. 
    
    When $f$ is convex (or concave) and $m_f=2^n$ for some $n \geqslant 1$, then there exists a neural network of the form \eqref{eq:def_discriminators} verifying Assumption \ref{ass:compactness}  that represents $f$. Its depth is $\log_2(m_f)+1$ and its size is at most $2m_f-1$.
\end{proposition}
This proposition is the counterpart of \citet[][Theorem 2.2]{arora2018understanding}, which states that any $m_f$-piecewise linear function from $\mathds{R} \to \mathds{R}$ can be represented by a $2$-layer ReLU neural network with a size at least $m_f-1$ . \citet[][Theorem 5.2]{he2018relu} shows that the upper-bound on the size of ReLU networks is $O(2^{m^2+2(m-1)})$. Thus, for the representation of piecewise linear functions on the real line, GroupSort networks require larger depths but smaller sizes. Besides, bear in mind that the obtained ReLU neural networks do not necessarily verify a requirement similar to the one of Assumption \ref{ass:compactness}.

Regarding the number of linear regions of GroupSort networks on the real line, we have the following result:
\begin{lemma}\label{lem:number_of_pieces}
    Any neural network of the form \eqref{eq:def_discriminators} on the real line, with depth $q$ and widths $\nu_1, \hdots, \nu_{q-1}$, parameterizes a piecewise linear function with at most $2^{q-2} \times (\nu_1/2+1) \times \nu_2 \times \cdots \times \nu_{q-1}$ linear subdomains. 
\end{lemma}
We deduce from this lemma that for a neural network of the form \eqref{eq:def_discriminators} with depth $q \geqslant 2$ and constant width $\nu$, the maximum number of linear regions is $O(2^{q-3}\nu^{q-1})$. Similarly to ReLU networks \citep{montufar2014number, arora2018understanding}, the maximum number of linear regions for GroupSort networks with grouping size $2$ is also likely to grow polynomially in $\nu$ and exponentially in $q$. 

Our next corollary now illustrates the trade-off between depth and width for GroupSort neural networks. 
\begin{corollary}\label{cor:lower_bound_for_nn}
    Let $f \in \emph{Lip}_1(\mathds R)$ be an $m_f$-piecewise linear function. Then, any neural network of the form \eqref{eq:def_discriminators} verifying Assumption \ref{ass:compactness} and representing $f$ with a depth $q$, has a size $s$ at least $\frac{1}{2}(q-1) m_f^{1/(q-1)}$. 
\end{corollary}
The lower bound highlighted in Corollary \ref{cor:lower_bound_for_nn} is dependent on the depth $q$ of the neural network. By looking at the minimum of the function, we get that any neural network representing $f$ has a size $s \geqslant \frac{e \ln(m_f)}{2}$. Thus, merging this result with Proposition \ref{cor:func_r_to_r}, we have that for any $m_f$-piecewise linear function from $\mathds{R} \to \mathds{R}$, there exists a GroupSort network verifying Assumption \ref{ass:compactness} with a size $s$ satisfying
\begin{equation*}
    \frac{e \ln(m_f)}{2} \leqslant s \leqslant 3m_f^2-m_f-3.
\end{equation*}
We realize that this inequality is large but, up to our knowledge, this is first of this type for GroupSort neural networks.

\subsection{Approximating Lipschitz continuous functions on compact sets} \label{section:lipschitz_function}
Following our plan, we tackle in this subsection the task of approximating Lipschitz continuous functions on compact sets using GroupSort neural networks. The space of continuous functions on $[0,1]^d$ is equipped with the uniform norm
\begin{equation*}
    \|f-g\|_\infty = \underset{x \in [0,1]^d}{\max} \ |f(x)-g(x)|.
\end{equation*}
The main result of the section, and actually of the article, is that GroupSort neural networks are well suited for approximating functions in $\text{Lip}_1([0,1]^d)$.
\begin{theorem}\label{th:approximating_any_lipschitz_function}
    Let $\varepsilon>0$ and $d\geqslant2$, $f\in \emph{Lip}_1([0,1]^d)$. Then there exists a neural network $D$ of the form \eqref{eq:def_discriminators} verifying Assumption \ref{ass:compactness} such that $\|f-D\|_{\infty} \leqslant \varepsilon$. The depth of $D$ is $O(d^2\log_2(\frac{2\sqrt{d}}{\varepsilon}))$ and its size is $O((\frac{2\sqrt{d}}{\varepsilon})^{d^2})$. 
\end{theorem}
To the best of our knowledge, Theorem \ref{th:approximating_any_lipschitz_function} is the first one that provides an upper bound on the depth and size of neural networks, with constraints on the weights, that approximate Lipschitz continuous functions. 

As for the representation of piecewise linear functions, one can, for the sake of completeness, compare this bound with those previously found in the literature of ReLU neural networks. \citet{yarotsky2017error} establishes the density of ReLU networks in Sobolev spaces, using a different technique of proof. In particular, Theorem 1 of this paper states that for any $f\in \text{Lip}_1([0,1]^d)$ continuously differentiable, there exists a ReLU neural network approximating $f$ with precision $\varepsilon$, with depth at most $c(\ln(1/\varepsilon)+1)$ and size at most $c \varepsilon^{-d}(\ln(1/\varepsilon) + 1)$ (with a constant $c$ function of $d$). Comparing this result with our Theorem \ref{th:approximating_any_lipschitz_function}, we see that, with respect to $\varepsilon$, both depths are similar but ReLU networks are smaller in size. However, one has to keep in mind that both lines of proof largely differ. Besides, our formulation ensures that the approximator is also a $1$-Lipschitz function, a feature that cannot be guaranteed under the formulation of \citet{yarotsky2017error}.

It turns out however that our framework provides smaller neural networks as soon as $d=1$.
\begin{proposition}\label{cor:approximating_real_valued_lipschitz_function}
    Let $\varepsilon>0$ and $f\in \emph{Lip}_1([0,1])$. Then there exists a neural network $D$ of the form \eqref{eq:def_discriminators} verifying Assumption \ref{ass:compactness} such that $\|f-D\|_{\infty} \leqslant \varepsilon$.  The depth of $D$ is $2 \lceil \log_2 (1/\varepsilon) \rceil+1$ and its size is  $O((\frac{1}{\varepsilon})^2)$.
    
    Besides, if $f$ is assumed to be convex or concave, then the depth of $D$ is $\lceil \log_2(1/\varepsilon) \rceil+1$ and its size is $O(\frac{1}{\varepsilon})$.
\end{proposition}

\section{Impact of the grouping size}\label{sec:grouping_size_k}
\begin{table*}
\begin{center}
\begin{tabular}{|l|c|c|c|c|}
\cline{2-5}
\multicolumn{1}{l|}{\textbf{Methods}} & Up Depth & Up Size & Down Size & Reference \\
\hline
\multicolumn{5}{l|}{\textbf{Representing $m=k^n$-PWL functions in $\mathds{R}^d$ with a constant width $\nu$}}  \\
\hline
ReLU & $\lceil \log_2(d+1) \rceil+1$ & $O(d2^{m^2})$ & $O(m)$ & \citet{he2018relu} \\[0.1cm]
GroupSort $GS=k$ & $\lceil 2\log_k(m) \rceil +1$ & $\frac{m^2-1}{k-1}$ & $\frac{\nu \log_k(m)}{2\log_k(\nu)}$ & present article\\[0.1cm]
\hline
\multicolumn{5}{l|}{\textbf{Approximating 1-Lipschitz continuous functions in $[0,1]^d$}} \\
\hline
ReLU & $O(\ln(\frac{1}{\varepsilon}))$ & $O(\frac{\ln(1/\varepsilon)}{\varepsilon^d})$ & $\backslash$ & \citet{yarotsky2017error} \\[0.1cm]
GroupSort $GS=\lceil \frac{2 \sqrt{d}}{\varepsilon} \rceil$ & $O(d^2)$  & $O((\frac{2\sqrt{d}}{\varepsilon})^{d^2-1})$ & $\backslash$ & present article\\[0.1cm]
\hline
\multicolumn{5}{l|}{\textbf{Approximating 1-Lipschitz continuous functions in $[0,1]$}} \\
\hline
ReLU (PWL representation) & 2  & $O(2^{1/\varepsilon^2+2/\varepsilon})$ & $\backslash$ & \citet[][]{he2018relu} \\[0.1cm]
ReLU (different approach)& $O(\ln(\frac{1}{\varepsilon}))$ & $O(\frac{\ln(1/\varepsilon)}{\varepsilon})$ & $\backslash$ & \citet{yarotsky2017error} \\[0.1cm]
Adaptative ReLU & $6$ & $O(\frac{1}{\varepsilon \ln(1/\varepsilon)})$ & $\backslash$ & \citet{yarotsky2017error} \\[0.1cm]
GroupSort $GS=\lceil \frac{1}{\varepsilon} \rceil$ & $3$ & $O(\frac{1}{\varepsilon})$ & $\backslash$ & present article\\
\hline
\end{tabular}
\end{center}
\caption{Summary of the results shown in the present paper together with results previously found for ReLU networks. ``Up Depth'' refers to upper bounds on the depths, ``Up Size'' to upper bounds on the sizes, and ``Down Size'' to lower bounds on the sizes. The symbol ``$\backslash$'' means that no result is known (up to our knowledge).\label{table:results}}
\end{table*}

The previous section paved the way for a better understanding of GroupSort neural networks and their ability to approximate Lipschitz continuous functions. As mentioned in Section \ref{section:background}, one can play with the grouping size $k$ of the neural network when defining its architecture. However, it is not clear how changing this parameter might influence the expressivity of the network. The present section aims at bringing some understanding. Following a similar reasoning as in Section \ref{sec:grouping_size_2}, we start by analyzing how GroupSort networks with an arbitrary grouping size $k\geqslant2$ can represent any piecewise linear functions: 
\begin{proposition}[Extension of Proposition \ref{lem:auxiliary_lemma_max_function}]\label{lem:auxiliary_lemma_max_function_GSk}
    Let $f_1, \hdots, f_m: \mathds{R}^d \to \mathds{R}$ be a collection of functions ($m \geqslant 2$), each represented by a neural network of the form \eqref{eq:def_discriminators}, with common depth $q$ and sizes $s_i$,  $i=1, \hdots, m$. 
    
    In the specific case where $m=k^n$ for some $n \geqslant 1$, there exist neural networks of the form \eqref{eq:def_discriminators} (with grouping size $k$) with depth $q+\log_k(m)$ and size at most $s_1 + \cdots + s_{m} + \frac{m -1}{k-1} -1$ that represent the functions $f= \max(f_1, \hdots, f_{m})$ and $g= \min(f_1, \hdots, f_{m})$. 
\end{proposition}
Similarly to Section \ref{sec:grouping_size_2}, this leads to the following corollary:
\begin{corollary}[Extension of Corollary \ref{cor:piecewise_linear_functions_on_convex_sets}] \label{cor:piecewise_linear_functions_on_convex_sets_arbitrary_k}
Let $f \in \emph{Lip}_1(\mathds R^d)$ that is also $m_f$-piecewise linear with convex subdomains $\Omega_1, \hdots, \Omega_{m_f}$ such that $m_f=k^n$ for some $n\geqslant1$. Then there exists a neural network of the form \eqref{eq:def_discriminators} verifying Assumption \ref{ass:compactness}  that represents $f$. Besides, its depth is $2\lceil \log_k(m_f) \rceil+1$ and its size is at most $\frac{m_f^2-1}{k-1}$. 
\end{corollary}
Proposition \ref{lem:auxiliary_lemma_max_function_GSk} and Corollary \ref{cor:piecewise_linear_functions_on_convex_sets_arbitrary_k} exhibit the nice properties of using larger grouping sizes. Indeed, for a given $q\geqslant1$, there exists a neural network with depth $2q+1$ and grouping size $k$ representing a function with $k^q$ pieces. Consequently, the use of larger grouping sizes helps have more expressive neural networks. The efficiency of larger grouping sizes may also be explained by the following result for GroupSort networks on the real line:
\begin{lemma}[Extension of Lemma \ref{lem:number_of_pieces}]\label{lem:number_of_pieces__GSk}
    Any neural network of the form \eqref{eq:def_discriminators} on the real line, with depth $q$, widths $\nu_1, \hdots, \nu_{q-1}$, and grouping size $k$, parameterizes a piecewise linear function with at most $k^{q-2} \times (\frac{(k-1)\nu_1}{2}+1) \times \nu_2 \times \cdots \times \nu_{q-1}$ linear subdomains. 
\end{lemma}
Thus, the number of linear regions of a GroupSort network is likely to increase polynomially with the grouping size, which highlights the benefits of using larger groups. Similarly to Section \ref{sec:grouping_size_2}, when moving to the approximation of Lipschitz continuous functions on $[0,1]^d$, we are lead to the following theorem:
\begin{theorem}[Extension Theorem \ref{th:approximating_any_lipschitz_function}] \label{th:approximating_any_lipschitz_function_arbitrary_k}
Let $\varepsilon>0$, $d\geqslant2$, and $f\in \emph{Lip}_1([0,1]^d)$. Then there exists a neural network $D$ of the form \eqref{eq:def_discriminators} verifying Assumption \ref{ass:compactness} with grouping size $\lceil \frac{2 \sqrt{d}}{\varepsilon} \rceil$ such that $\|f-D\|_{\infty} \leqslant \varepsilon$. The depth of $D$ is $O(d^2)$ and its size is $O((\frac{2\sqrt{d}}{\varepsilon})^{d^2-1})$. 
\end{theorem}
Using a grouping size proportional to $1/\varepsilon$, we thus have a bound on the depth that is independent from the error rate. The uni-dimensional case leads to a different result:
\begin{proposition}[Extension of Proposition \ref{cor:approximating_real_valued_lipschitz_function}]\label{cor:approximating_real_valued_lipschitz_function_arbitrary_k}
    Let $\varepsilon>0$ and $f\in \emph{Lip}_1([0,1])$. Then there exists a neural network $D$ of the form \eqref{eq:def_discriminators} verifying Assumption \ref{ass:compactness} (with grouping size $k$) such that $\|f-D\|_{\infty} \leqslant \varepsilon$. The depth of $D$ is $2\lceil \log_k(\frac{1}{\varepsilon}) \rceil+1$ and its size is at most $O(\frac{1}{k\varepsilon^2})$.
    
    In particular, if $k$ is chosen to be equal to $\lceil \frac{1}{\varepsilon} \rceil$, then the depth of $D$ is $3$ and its size is $O(\frac{1}{\varepsilon})$. 
\end{proposition}
When approximating real-valued functions, the use of larger grouping sizes can significantly decrease the required size since it goes from $O(1/\varepsilon^2)$ in Proposition \ref{cor:approximating_real_valued_lipschitz_function} to $O(1/\varepsilon)$ in Proposition \ref{cor:approximating_real_valued_lipschitz_function_arbitrary_k}. When $f$ is assumed to be convex or concave, the depth of the network $D$ can further be reduced to $2$. 


Using a different approach for approximating Lipschitz continuous functions in $[0,1]$, \citet[][Theorem 1]{yarotsky2017error} shows that ReLU networks with a depth of $O(\ln(1/\varepsilon))$ is needed together with a size $O(\frac{\ln(1/\varepsilon)}{\varepsilon})$ to approximate with an error rate $\varepsilon$. To sum-up, when compared with ReLU networks, GroupSort neural networks with well-chosen grouping size can be significantly more expressive.

Table \ref{table:results} summarizes the results shown in the present paper together with results previously found for ReLU networks. Bear in mind that GroupSort neural networks also have the supplementary condition that any parameterized function verifies the $1$-Lipschitz continuity.

\section{Experiments}\label{section:experiments}
\citet{Anil2018SortingOL} have already compared the performances of GroupSort neural networks with their ReLU counterparts, both with constraints on the weights. In particular, they showed that ReLU neural networks are more sensitive to adversarial attacks while stressing the fact that if their weights are limited, then these networks lose their expressive power. Building on these observations, we further illustrate the good behavior of GroupSort neural networks in the context of estimating a Lipschitz continuous regression function and in approximating the Wasserstein distance (via its dual form) between pairs of distributions.

\begin{figure}[h]
    \centering
    \subfloat[$q=2$]
    {
        \includegraphics[width=0.47\linewidth]{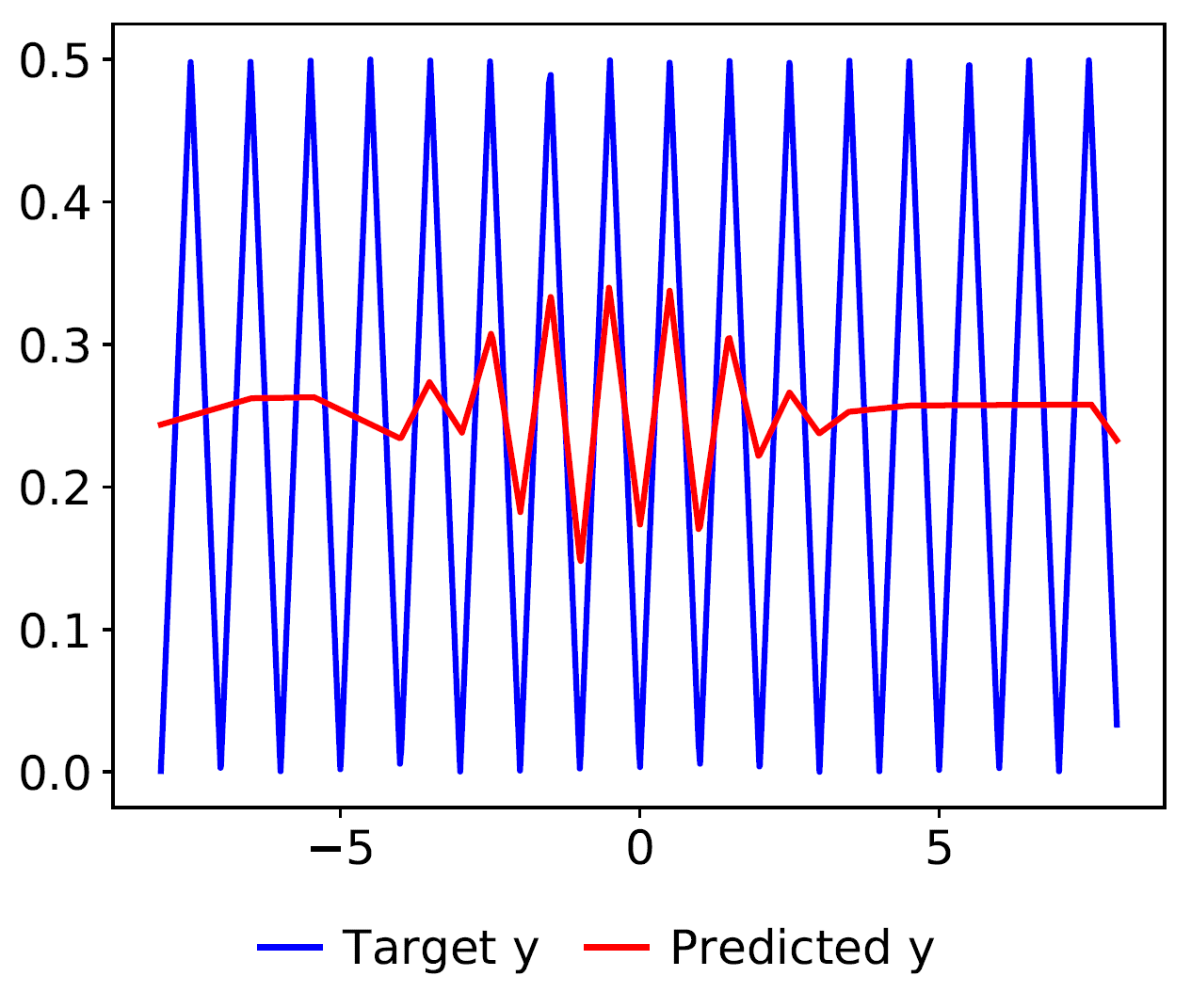}
    }
    \subfloat[$q=20$]
    {
        \includegraphics[width=0.47\linewidth]{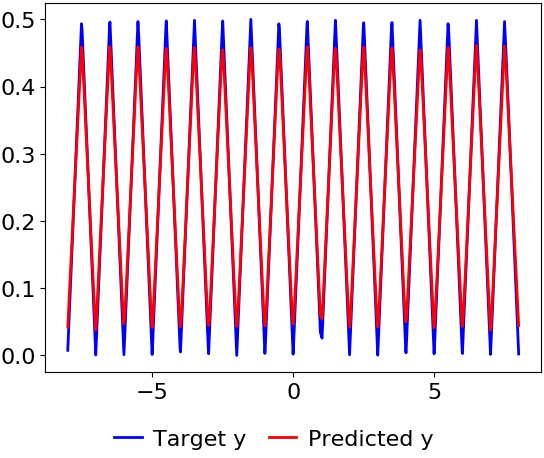}
    }\\
    \subfloat[Uniform norm]
    {
        \includegraphics[width=0.47\linewidth]{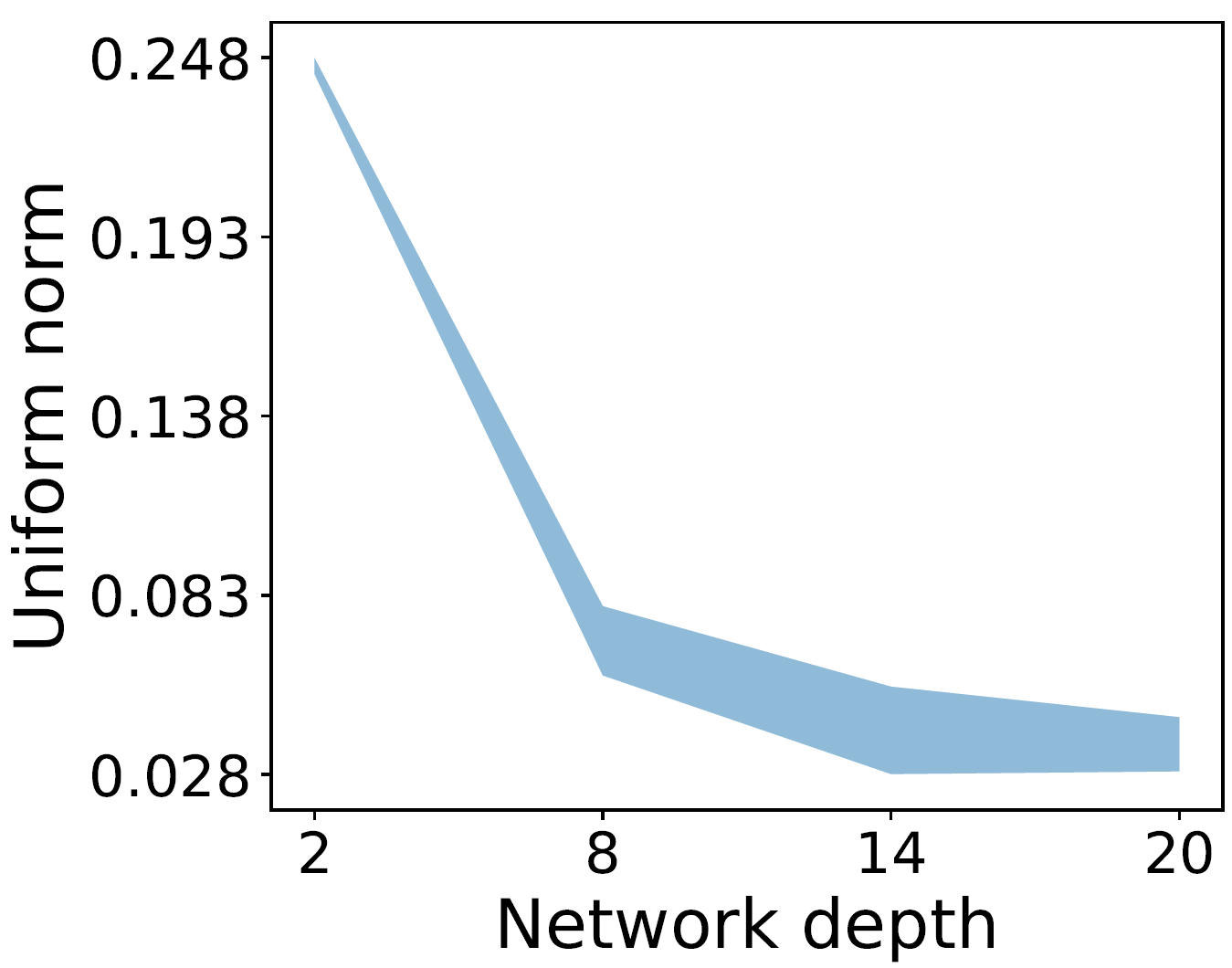}
    }
    \subfloat[Lipschitz constants]
    {
        \includegraphics[width=0.47\linewidth]{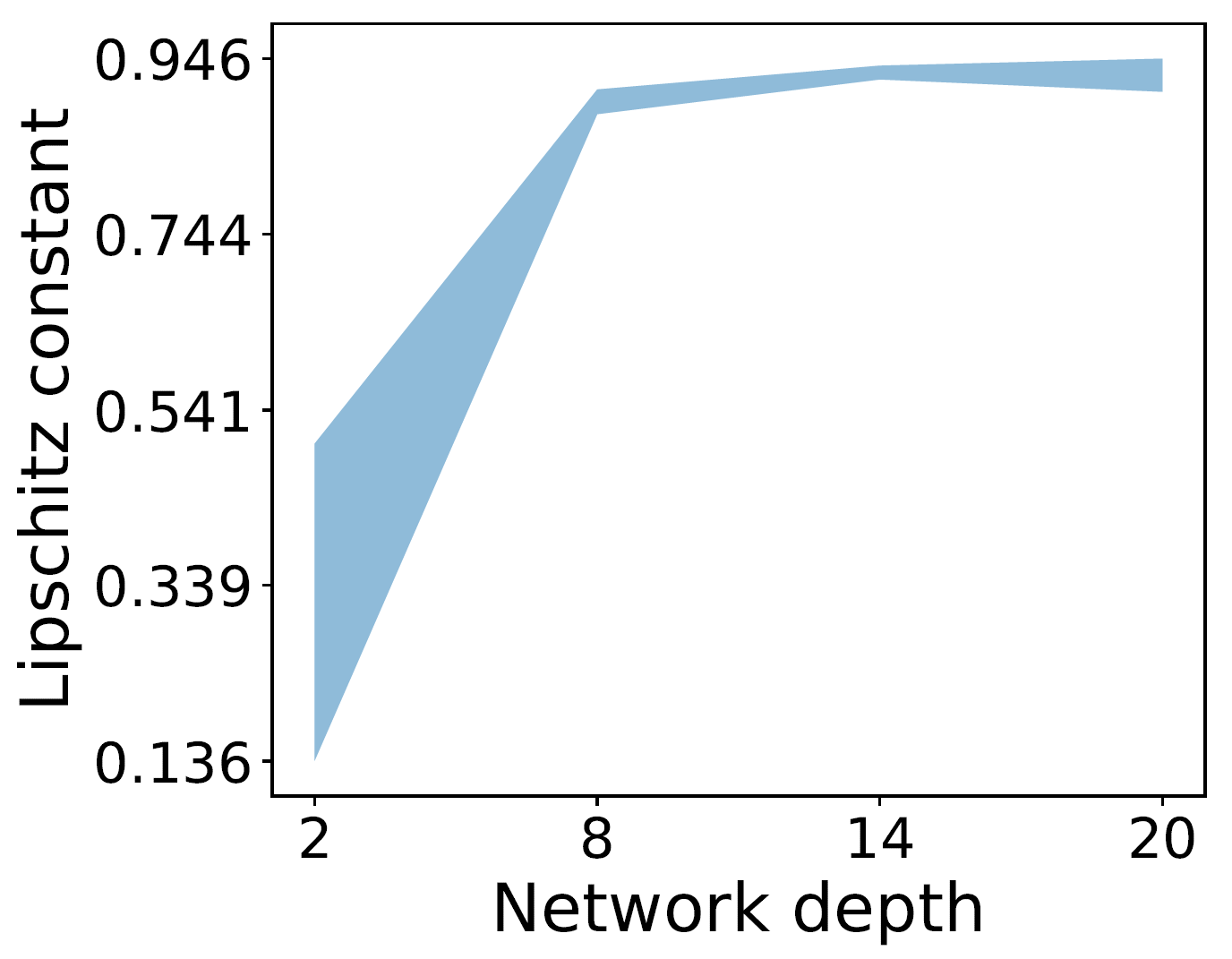}
    }
    \caption{Reconstruction of a $32$-piecewise linear function on $[-8,8]$ with a GroupSort neural network of the form \eqref{eq:def_discriminators} with depth $q = 2$, $8$, $14$, $20$, and a constant width $\nu=50$ (the thickness of the line represents a $95\%$-confidence interval).}
    \label{fig:comparing_results_pwl}
\end{figure}

\paragraph{Impact of the depth.}
We start with the problem of learning a function $f$ in the model $Y=f(X)$, where $X$ follows a uniform distribution on $[-8, 8]$ and $f$ is $32$-piecewise linear. To this aim, we use neural networks of the form \eqref{eq:def_discriminators} with respective depth $q = 2$, $8$, $14$, $20$, and a constant width $\nu=50$. Since we are only interested in the approximation properties of the networks, we assume to have at hand an infinite number of pairs $(X_i,f(X_i))$ and train the models by minimizing the mean squared error. We give in the Appendix, the full details of our experimental setting. The quality of the estimation is evaluated using the uniform norm between the target function $f$ and the output network. In order to enforce Assumption \ref{ass:compactness}, GroupSort neural networks are constrained using the orthonormalization of \citet{bjorck1971iterative}. The results are presented in Figure \ref{fig:comparing_results_pwl}. Note that throughout this section, confidence intervals are computed over 20 runs. In line with Theorem \ref{th:disc_can_represent_piecewise_linear}, which states that $f$ is representable by a neural network of the form \eqref{eq:def_discriminators} with size at most $3\times32^2 + 32 -1 = 3104$, we clearly observe that, as the depth of the networks increases, the uniform norm decreases and the Lipschitz constant of the network converges to 1. The reconstruction of this piecewise linear function is even almost perfect for the depth $q=20$, i.e., with a network of size only $20 \times 60=1200$, a value significantly smaller than the upper bound of the theorem. 

\begin{figure}[h]
    \centering
    \subfloat[$q=2$]
    {
        \includegraphics[width=0.47\linewidth]{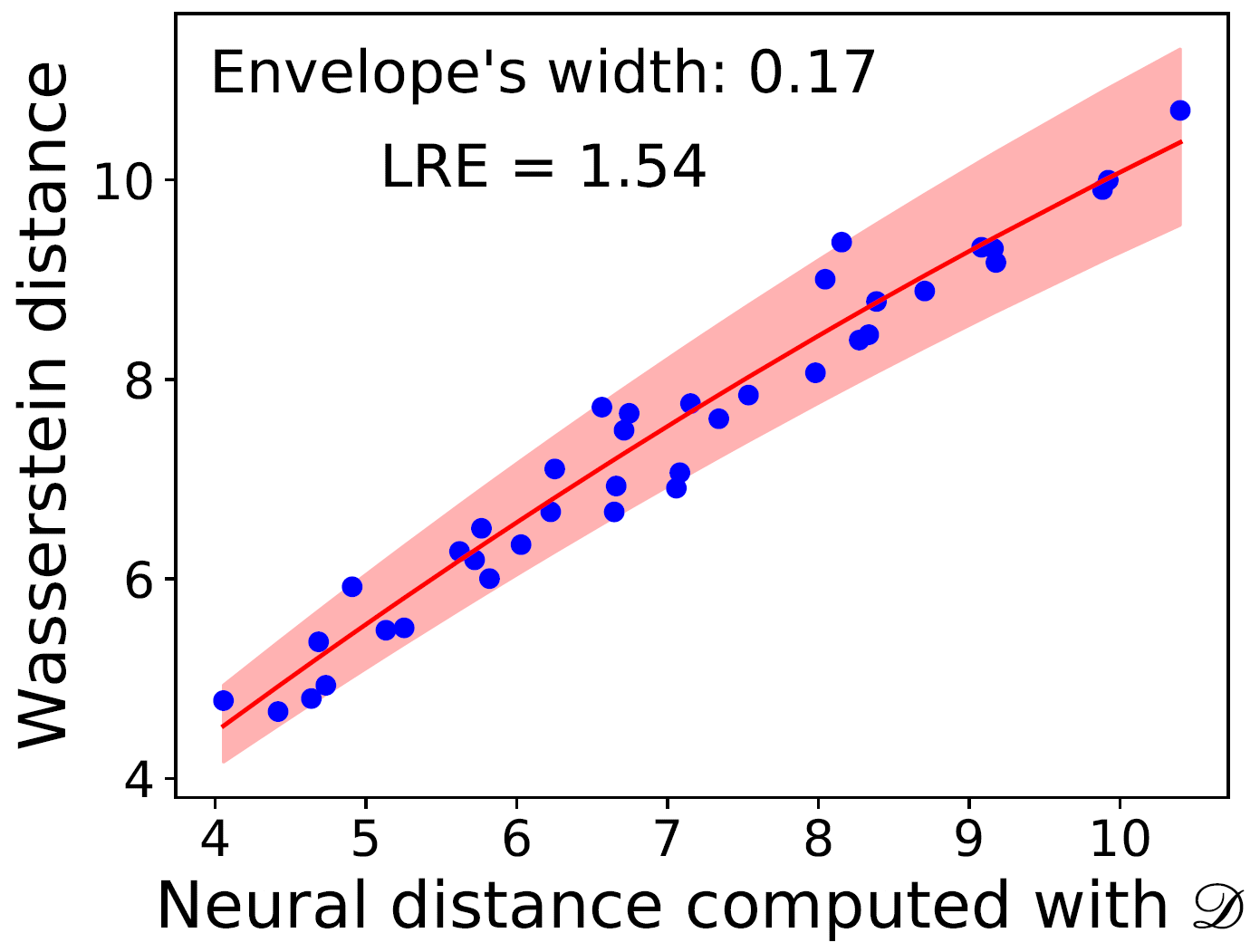}
    }
    \subfloat[$q=5$]
    {
        \includegraphics[width=0.47\linewidth]{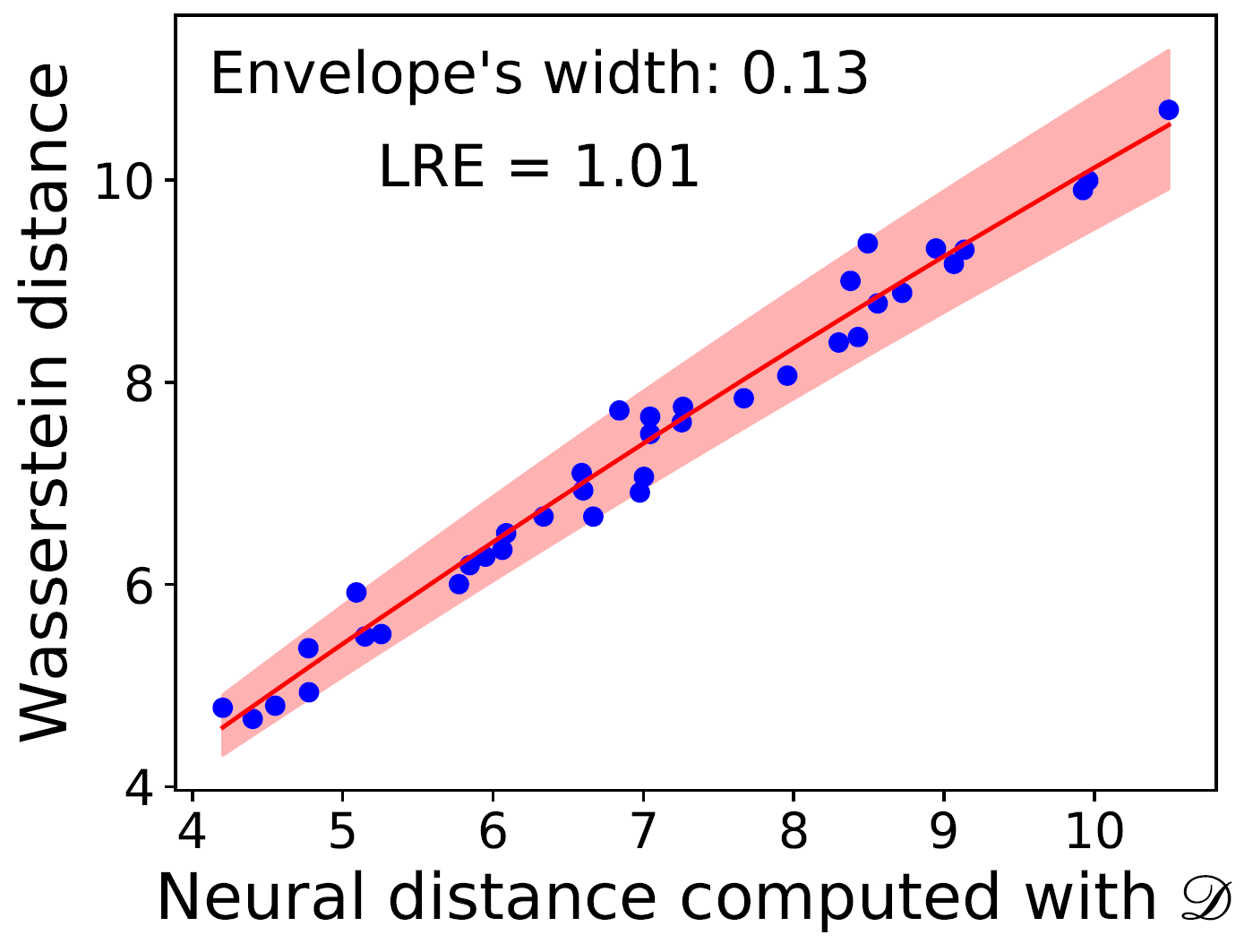}
    }
    \caption{Scatter plots of $40$ pairs of Wasserstein and neural distances computed with GroupSort neural networks, for $q=2, 5$. The underlying distributions are bivariate Gaussians. The red curve is the optimal parabolic fitting and LRE refers to the Least Relative Error. The red zone is the envelope obtained by stretching the optimal curve.}
 \label{fig:calculate_neural_dists}
\end{figure}

We also illustrate the behavior of GroupSort neural networks in the context of WGANs \citep{arjovsky2017wasserstein}. We run a series of small experiments in the simplified setting where we try to approximate the $1$-Wasserstein distance between two bivariate mixtures of independent Gaussian distributions with $4$ components. We consider networks of the form \eqref{eq:def_discriminators} with grouping size $2$, a depth $q=2$ and $q=5$, and a constant width $\nu=20$. 
For a pair of distributions $(\mu, \nu)$, our goal is to exemplify the relationship between the $1$-Wasserstein distance $\sup_{f \in \text{Lip}_1(\mathds{R} ^2)} (\mathds{E}_\mu - \mathds{E}_\nu)$ \citep[approximated with the Python package by][]{flamary2017pot} and the neural distance $\sup_{f \in \mathscr{D}_2} (\mathds{E}_\mu - \mathds{E}_\nu$) \citep{Arora0LMZ17} computed over the class of functions $\mathscr{D}_2$. To this aim, we randomly draw $40$ different pairs of distributions. Then, for each of these pairs, we compute an approximation of the $1$-Wasserstein distance and calculate the corresponding neural distance. Figure \ref{fig:calculate_neural_dists} depicts the best parabolic fit between $1$-Wasserstein and neural distances, and shows the corresponding Least Relative Error (LRE) together with the width of the envelope. The take-home message of this figure is that both the LRE and the width are significantly smaller for deeper GroupSort neural networks. 

\begin{figure}[h]
    \centering
    \subfloat[Grouping size = 2\label{fig:GS2}]
    {
        \includegraphics[width=0.475\linewidth]{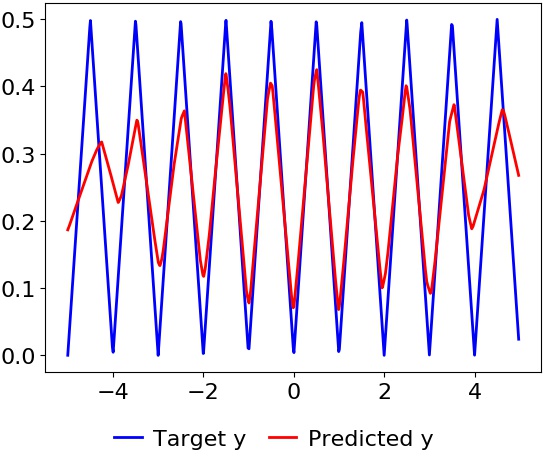}
    }
    \subfloat[Grouping size = 6\label{fig:GS5}]
    {
        \includegraphics[width=0.475\linewidth]{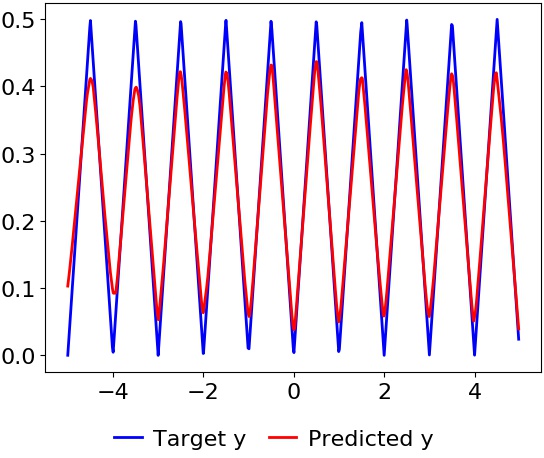}
    }\\
    \subfloat[Uniform norm\label{fig:GS_uniform_norm}]
    {
        \includegraphics[width=0.475\linewidth]{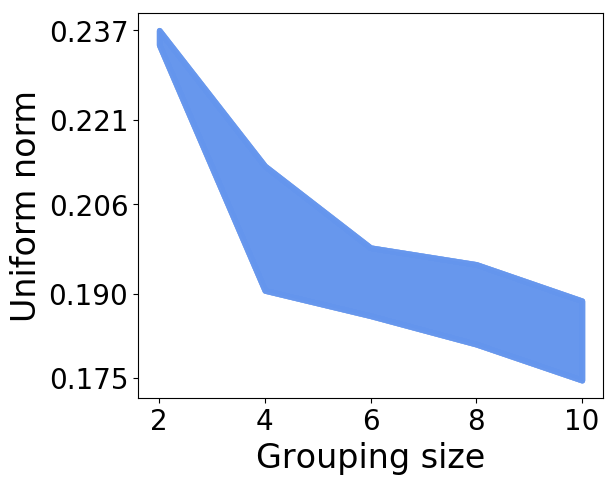}
    }
    \subfloat[Linear regions\label{fig:GS_linear_regions}]
    {
        \includegraphics[width=0.475\linewidth]{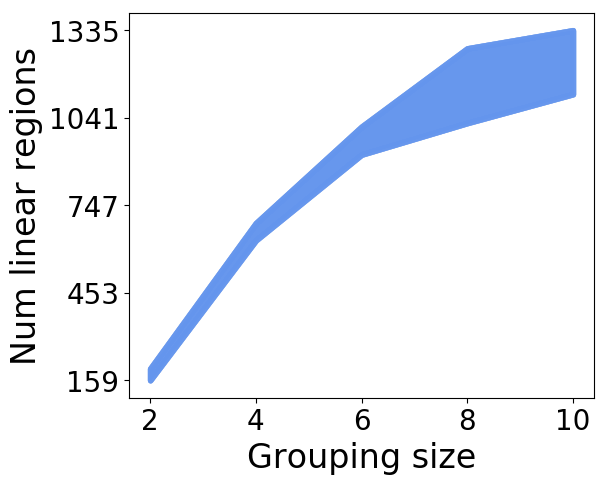}
    }
    \caption{Reconstruction of a $20$-piecewise linear function on $[-5,5]$ (top line) and a $64$-piecewise linear function (bottom line) with GroupSort neural networks of the form \eqref{eq:def_discriminators} with depth $q=4$ and varying grouping sizes $k=2, 4, 6, 8, 10$.}
    \label{fig:grouping_size}
\end{figure}

\paragraph{Impact of the grouping size.}
To highlight the benefits of using larger grouping sizes, we show the impact of increasing the grouping size from $2$ in Figure \ref{fig:GS2} to $5$ in Figure \ref{fig:GS5} for the representation of a $20$-piecewise linear function. This is corroborated by Figure \ref{fig:GS_uniform_norm}, which illustrates that the uniform norm with a $64$-piecewise linear function decreases when the grouping size increases. As already underlined in Lemma \ref{lem:number_of_pieces__GSk}, this may be explained by the fact that the number of linear regions significantly grows with the grouping size---see Figure \ref{fig:GS_linear_regions}.

\begin{figure*}
    \centering
    \subfloat[Prediction quality \label{fig:4a}]
    {
        \includegraphics[width=0.30\linewidth]{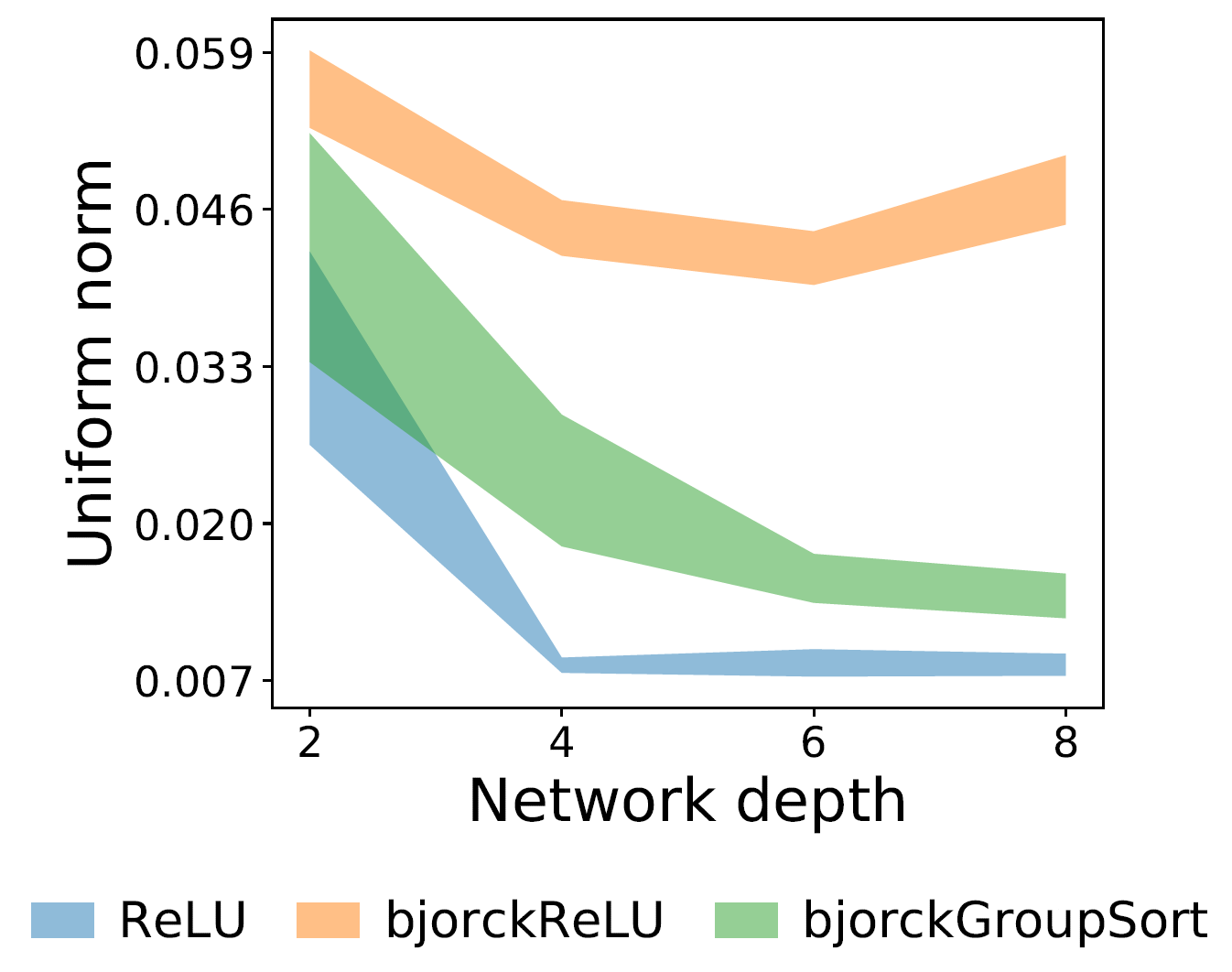}
    }
    \hfill
    \subfloat[Lipschitz constants \label{fig:4b}]
    {
        \includegraphics[width=0.30\linewidth]{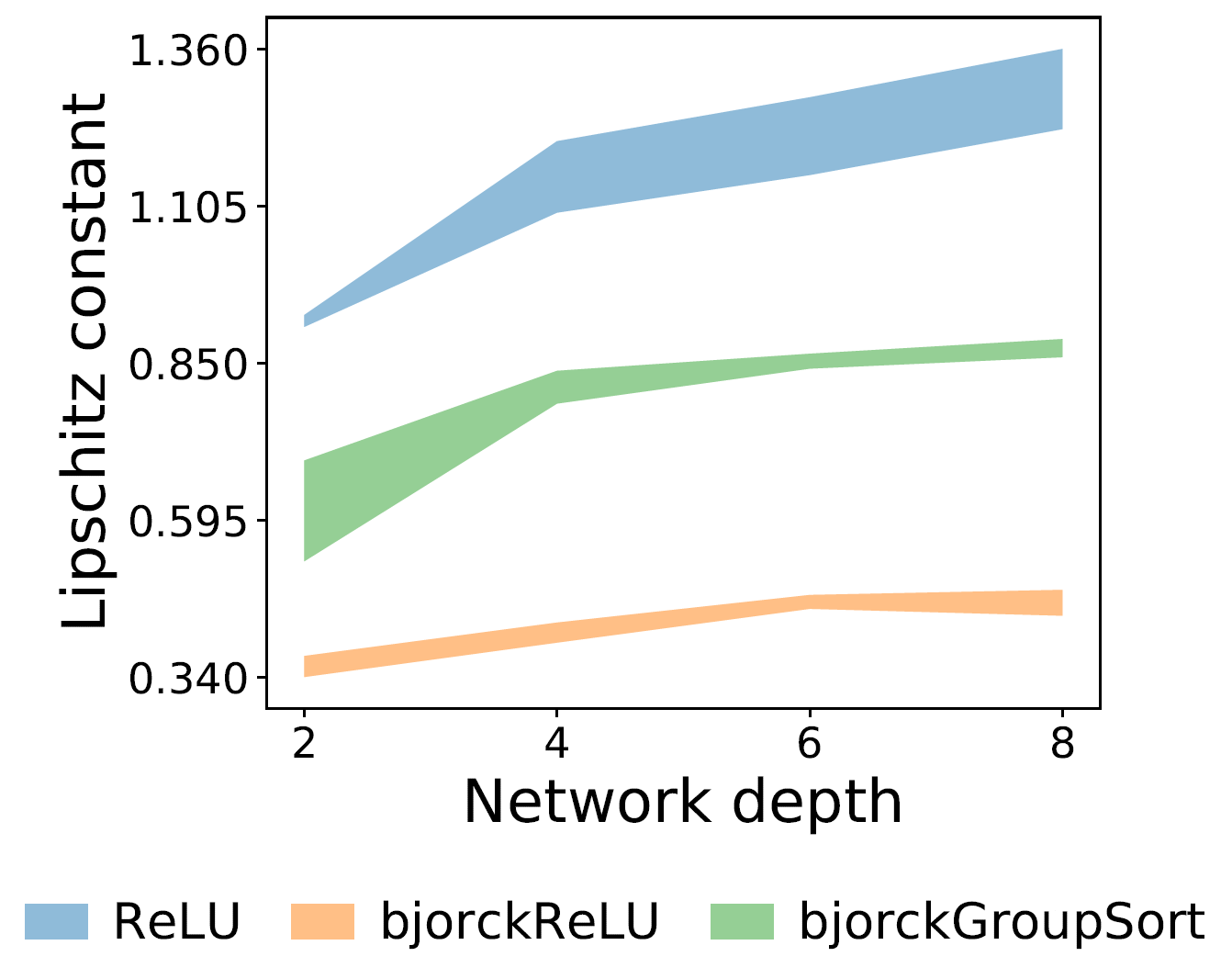}
    }\hfill
    \subfloat[Number of linear regions \label{fig:4c}]
    {
        \includegraphics[width=0.30\linewidth]{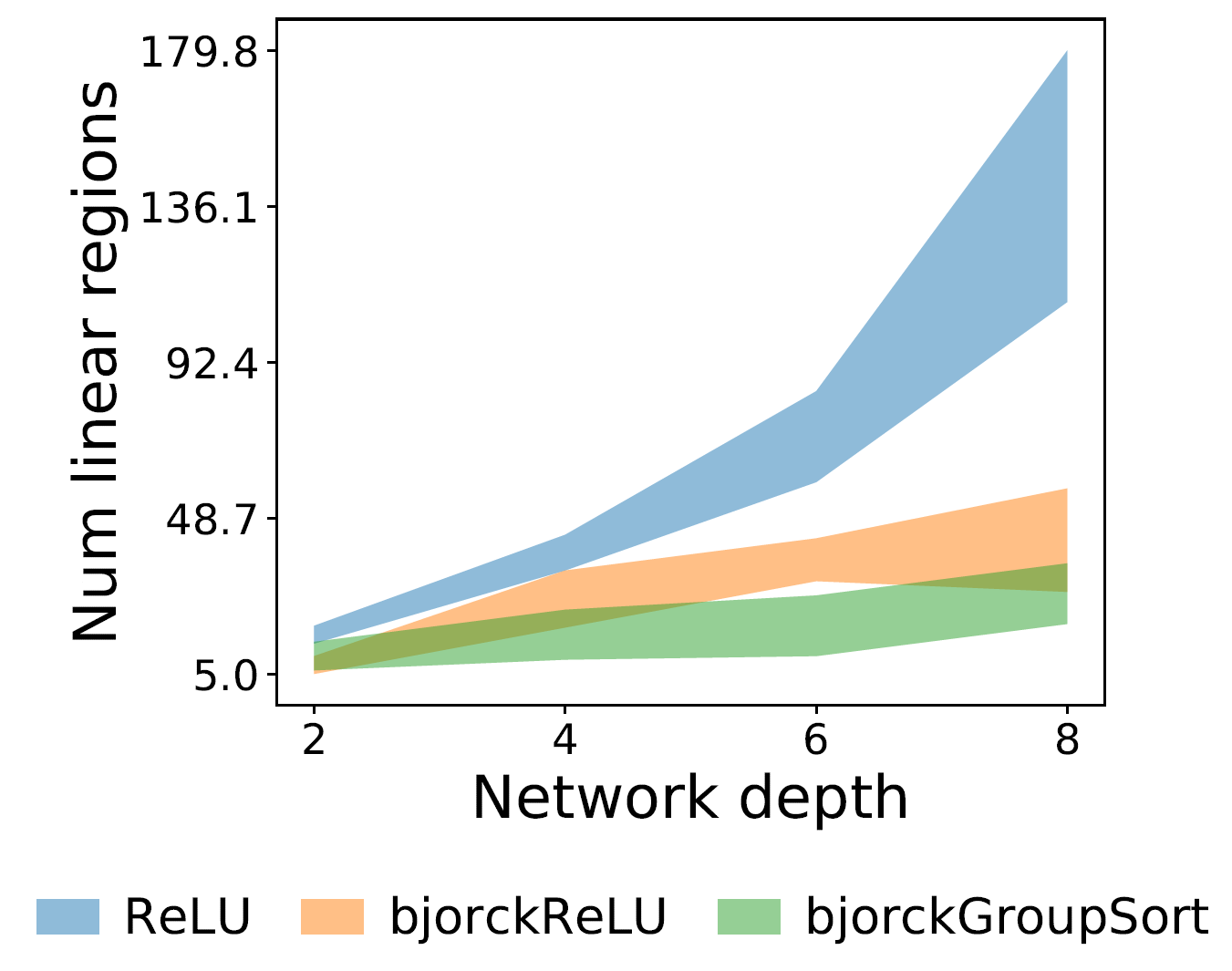}
    }
    \\
    \subfloat[Prediction quality]
    {
        \includegraphics[width=0.30\linewidth]{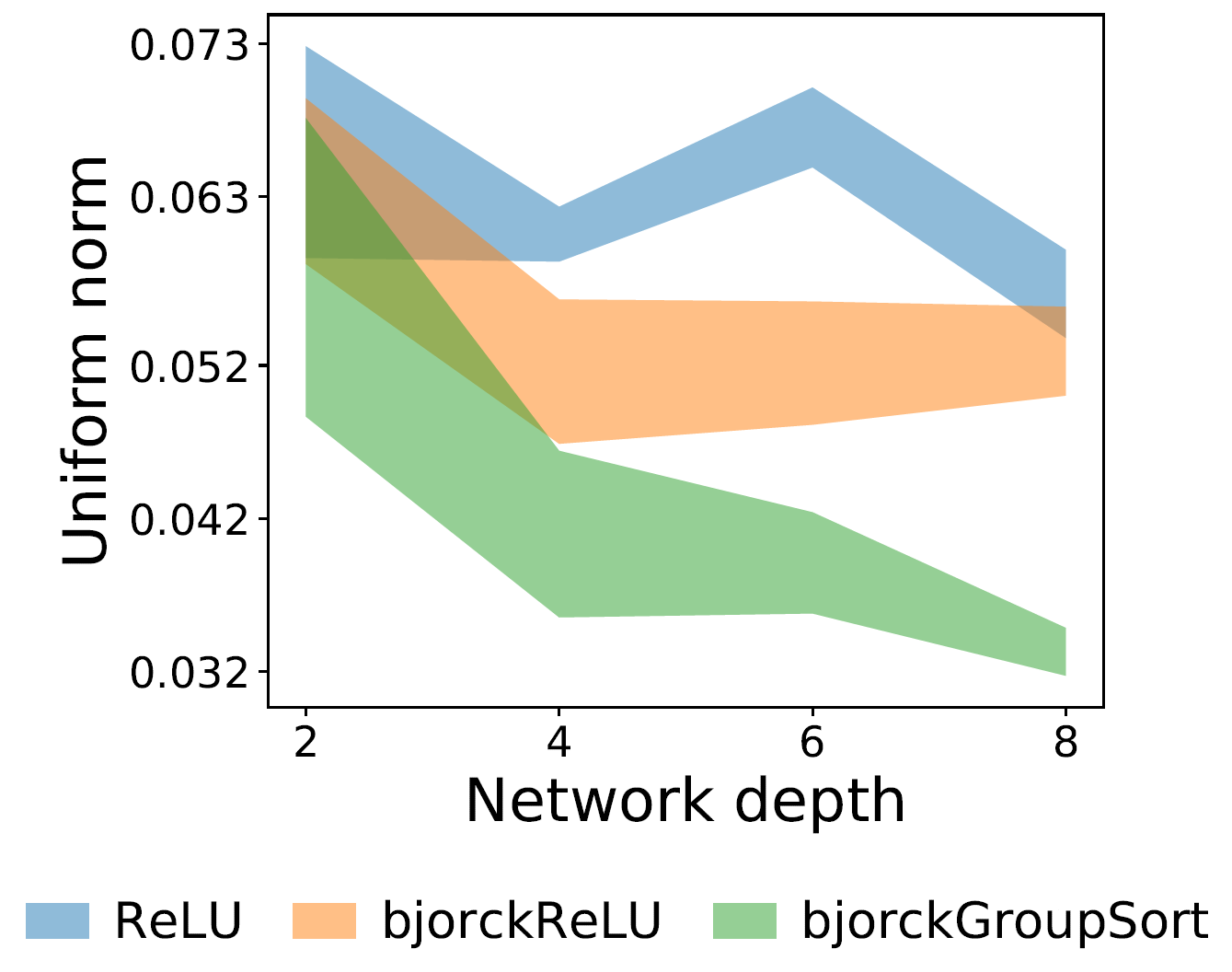}
    }\hfill
    \subfloat[Lipschitz constant\label{fig:4e}]
    {
        \includegraphics[width=0.30\linewidth]{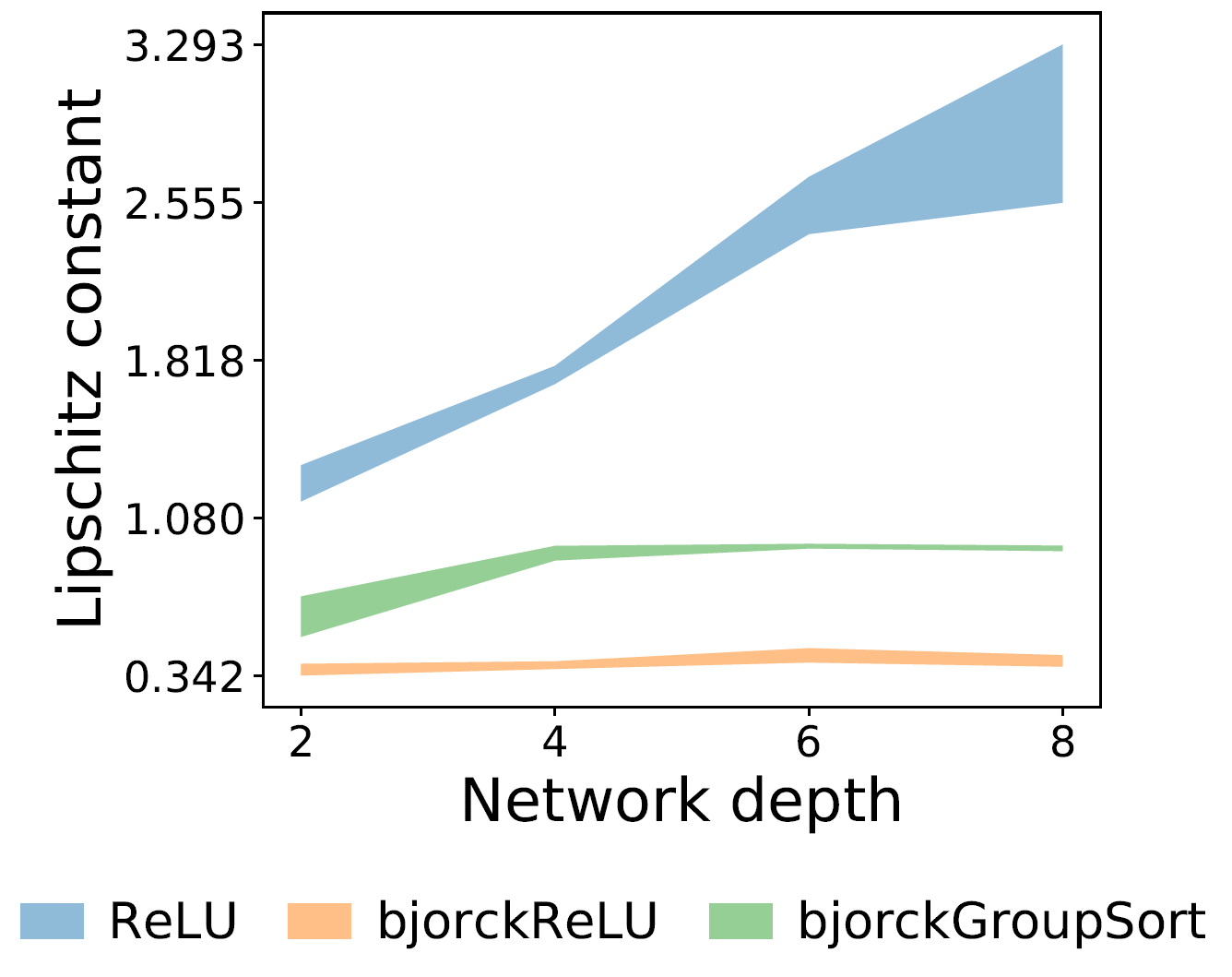}
    }\hfill
    \subfloat[Number of linear regions \label{fig:4f}]
    {
        \includegraphics[width=0.30\linewidth]{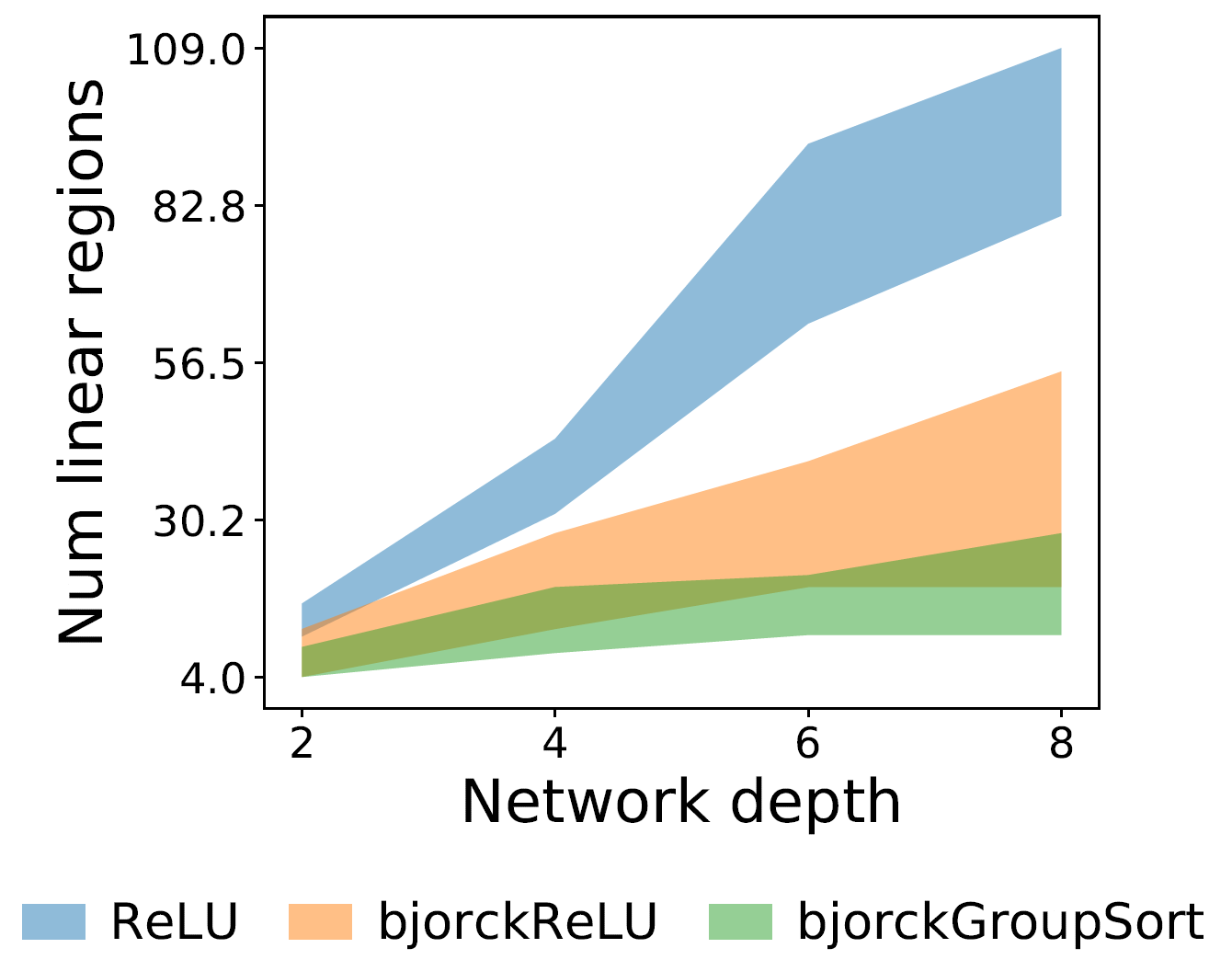}
    }
    \caption{(Top line) Estimating the function $f(x) = (1/15) \sin(15x)$ on $[0,1]$ in the model $Y=f(X)$, with a dataset of size $n=100$. 
    (Bottom line) Estimating the function $f(x) =(1/15) \sin(15x)$ on $[0,1]$ in the model $Y=f(X)+\varepsilon$, with a dataset of size $n=100$ (the thickness of the line represents a $95\%$-confidence interval).}
    \label{fig:nonoise_and_noise}
\end{figure*}

\paragraph{Comparison with ReLU neural networks.}
Next, in a second series of experiments, we compare the performances of GroupSort networks against two baselines: ReLU neural networks without constraints on the weights \citep[dense in the set of continuous functions on a compact set; see][]{yarotsky2017error}, and ReLU neural networks with orthonormalization of \citet{bjorck1971iterative}. The architecture of the ReLU neural networks in terms of depth and width is the same as for GroupSort networks: $q =2$, $4$, $6$ ,$8$, and $w=20$. The task is now to approximate the $1$-Lipschitz continuous function $f(x)= (1/15) \sin(15x)$ on $[0,1]$ in the models $Y=f(X)$ (noiseless case) and $Y=f(X)+\varepsilon$ (noisy case), where $X$ is uniformly distributed on $[0,1]$ and $\varepsilon$ follows a Gaussian distribution with standard deviation $0.05$. In both cases, we assume to have at hand a finite sample of size $n=100$ and fit the models by minimizing the mean squared error. 

Both results (noiseless case and noisy case) are presented in Figure \ref{fig:nonoise_and_noise}. We observe that in the noiseless setting Figure \ref{fig:4a}, \ref{fig:4b}, and \ref{fig:4c}, ReLU neural networks without normalization have a slightly better performance with respect to the uniform norm with, however, a Lipschitz constant larger than 1. On the other hand, in the noisy case, ReLU neural networks without constraints have a tendency to overfitting (a high Lipschitz constant close to $2.7$), leading to a deteriorated performance, contrary to GroupSort neural networks. Furthermore, in both cases (noiseless and noisy), ReLU with constraints are found to perform worse (due to a Lipschitz constant much smaller than $1$) than their GroupSort counterparts in terms of prediction. Interestingly, we see in the two examples shown in Figure \ref{fig:4e} and Figure \ref{fig:4f}, that the number of linear regions for GroupSort neural networks is smaller than for ReLU networks. 

Finally, we quickly show in Appendix a comparison between GroupSort and ReLU networks when approximating Wasserstein distances. The take home message is that, on this specific task, GroupSort networks perform better.

\section{Conclusion}
The results presented in this article show the advantage of using GroupSort neural networks over standard ReLU networks. On the one hand, ReLU neural networks without any constraints are sensitive to adversarial attacks (as they may have a large Lipschitz constant) and, on the other hand, lose expressive power when enforcing limits on their weights. On the opposite, GroupSort neural networks with constrained weights are proved to be both robust and expressive, and are therefore an interesting alternative. Moreover, by allowing larger grouping sizes for GroupSort networks, one can further increase their expressivity. These properties open new perspectives for broader use of GroupSort networks. 


\bibliography{main}

\appendix
\onecolumn
\section{Technical results and complementary experiments}
\subsection{Proof of Lemma \ref{lem:uniformly_lipschitz_neural_nets}}
We prove the result for $\mathcal{D}_2$. The result for $\mathcal{D}_k$ holds following a similar argument.

Fix $D_{2,\alpha} \in \mathscr D_2$, $\alpha \in \Lambda$. According to \eqref{eq:def_discriminators}, we have, for $x\in \mathds{R}^d$, $D_{2,\alpha}(x) = f_q \circ \cdots \circ f_1(x)$, where $f_i(t)= \sigma_2(V_it + c_i)$ for $i=1, \hdots, q-1$ ($\sigma_2$ is applied on pairs of components), and $f_{q}(t)=V_qt+c_q$. Therefore, for $(x,y) \in (\mathds{R}^d)^2$,
    \begin{align*}
        \|f_1(x) - f_1(y) \|_\infty & \leqslant \| V_1 x - V_1 y \|_\infty \\
        & \quad \mbox{(since $\sigma_2$ is $1$-Lipschitz)}\\
        &= \|V_1(x-y) \|_{\infty} \\
        & \leqslant \|V_1\|_{2, \infty} \ \|x-y\| \\
        &\leqslant \|x-y\| \\
        & \quad \mbox{(by Assumption \ref{ass:compactness})}.
    \end{align*}
    Thus,
 \begin{align*}
        \|f_2 \circ f_1(x) - f_2 \circ f_1(y) \|_\infty &\leqslant \| V_2f_1(x)  - V_2 f_1(y) \|_\infty \\
        & \quad \mbox{(since $\sigma_2$ is $1$-Lipschitz)}\\
        & \leqslant \|V_2\|_\infty \ \|f_1(x)-f_1(y)\|_\infty \\
        &\leqslant \|f_1(x)-f_1(y)\|_\infty \\
        & \quad \mbox{(by Assumption \ref{ass:compactness})}\\
        & \leqslant \|x-y\|.
    \end{align*}
Repeating this, we conclude that, for each $\alpha \in \Lambda$ and all $(x, y) \in (\mathds{R}^d)^2$, $|D_{2,\alpha}(x) - D_{2,\alpha}(y)| \leqslant \|x-y\|$, which is the desired result.

\subsection{Proof of Lemma \ref{lem:number_of_ordered_subdomains}}
Recall that $m_f \geqslant 2$. Throughout the proof, we let $\cdot$ refer to the dot product in $\mathds{R}^d$. Let $(i,j) \in \{1, \hdots, m_f\}^2$, $i \neq j$. There exist $(a_i, b_i) \in \mathds{R}^d \times \mathds{R}$ and $(a_j, b_j) \in \mathds{R}^d \times \mathds{R}$ such that $\ell_i = a_i \cdot x + b_i$ and $\ell_j = a_j \cdot x + b_j$. Therefore, 
\begin{equation*}
\ell_i(x) - \ell_j(x) \leqslant 0  \iff x \cdot (a_i-a_j) \leqslant b_j - b_i.
\end{equation*}
So, there exist two subdomains $\Tilde{\Omega}_1$ and $\Tilde{\Omega}_2$, separated by an affine hyperplane, in which $\ell_i - \ell_j$ does not change sign. By repeating this operation for the $m_f(m_f-1)/2$ different pairs $(\ell_i, \ell_j)$, we get that the number $M_f$ of subdomains on which any pair $\ell_i-\ell_j$ does not change sign is smaller than the maximal number of arrangements of $m_f(m_f-1)/2$ hyperplanes. 

Denoting by $C_{n,d}$ the maximal number of arrangements of $n$ hyperplanes in $\mathds{R}^d$, we know that when $d>n$ then $C_{n,d} = 2^n$, whereas if $n>d$ the upper bound $C_{n,d} \leqslant (1+n)^d$ becomes preferable \citep[][Chapter 30]{devroye2013probabilistic}. Thus, we have
\begin{equation*}
    m_f \leqslant M_f \leqslant \min\big(2^{m_f^2/2}, (m_f/\sqrt{2})^{2d}\big).
\end{equation*}

\subsection{Proof of Proposition \ref{lem:auxiliary_lemma_max_function}}
We prove the first part of the proposition by using an induction on $n$. The case where $n=1$ and thus $m=2^1$ is clear since the function $f = \max(f_1, f_2)$ can be represented by a neural network of the form \eqref{eq:def_discriminators} with depth $q+1$ and size $s_1+s_2+1$. Now, let $m=2^n$ with $n > 1$. We have that $m/2 = 2^{n-1}$. By the induction hypothesis, $g_1=\max(f_1, \hdots, f_{m/2})$ and $g_2=\max(f_{m/2+1}, \hdots, f_m)$ can be represented by neural networks of the form \eqref{eq:def_discriminators} with depths $q + n-1$, and sizes at most $s_1 + \cdots + s_{m/2} +  m/2 -1$ and $s_{m/2 + 1} + \cdots + s_m + m/2 -1 $, respectively. Consequently, the function $G(x)= (g_1(x),g_2(x))$ can be implemented by a neural network of the form \eqref{eq:def_discriminators} with depth $q+ n-1$ and size $s_1+\cdots+s_m+m-2$. Finally, by concatenating a one neuron layer, we have that the function $f= \max(g_1, g_2)$ can be represented by a neural network of the form \eqref{eq:def_discriminators} with depth $q+ n = q+ \log_2(m)$ and size at most $s_1+\cdots+s_m+m-1$.

Now, let us prove the case where $m$ is arbitrary. Let $f_1, \hdots, f_m: \mathds{R}^d \to \mathds{R}$ be a collection of functions ($m \geqslant 2$), each represented by a neural network of the form \eqref{eq:def_discriminators} with depth $q$ and size $s_i$,  $i=1, \hdots, m$. We prove below by an induction on $n$ that there exists a neural network of the form \eqref{eq:def_discriminators} with depth $q + \lceil\log_2(m) \rceil$, a final layer of width $\nu_{q-1}=2$, and a size at most $s_1 + \cdots + s_{m} + 2^{\lceil\log_2(m) \rceil} -1$ that represents the functions $f= \max(f_1, \hdots, f_{m})$ and $g= \min(f_1, \hdots, f_{m})$ (the symbol $\lceil \cdot \rceil$ stands for the ceiling function and the symbol $\lfloor \cdot \rfloor$ stands for the integer function). 

The base case $m=2$ is clear using the GroupSort activation and $\nu_1=2$. For $m > 2$, let $n\geqslant 2$ be such that $2^{n-1} \leqslant m < 2^n$. Let $g_1 =  \max(f_1, \hdots,f_{ 2^{n-1} })$ and $g_2 = \max(f_{2^{n-1}+1}, \hdots ,f_m)$. From the first part of the proof, we know that $g_1$ can be represented by a neural network of the form \eqref{eq:def_discriminators} with depth $q_1 = q+ \lfloor \log_2 m \rfloor = q+n-1$ and size $s_1 + \cdots + s_{2^{n-1}} + 2^{n-1}-1$. Also, by the induction hypothesis, $g_2$ can be represented by a neural network of the form \eqref{eq:def_discriminators} with depth $q_2 = q + \lceil \log_2 (m - 2^{n-1}) \rceil$ and size at most $s_{{2^{n-1}+1}} + \cdots + s_m + 2^{\lceil \log_2 (m - 2^{n-1}) \rceil} -1$. Therefore, by padding identity matrices with two neurons (recall that $\nu_{q_2-1}=2$) on layers from $q+\lceil \log_2 (m - 2^{n-1}) \rceil$ to $q+n-1$, we have:
\begin{align*}
    2^{\lceil \log_2 (m - 2^{n-1}) \rceil} -1 + 2(n-2-\lceil \log_2 (m - 2^{n-1}) \rceil)  &= \sum_{k=0}^{k=\lceil \log_2 (m - 2^{n-1}) \rceil-1} 2^k + \sum_{k=\lceil \log_2 (m - 2^{n-1}) \rceil}^{k=n-2} 2^1 \\
    &\leqslant \sum_{k=0}^{k=n-2} 2^k = 2^{n-1} - 1.
\end{align*}
Thus, $g_2$ can be represented by a neural network of the form \eqref{eq:def_discriminators} with depth $q_2 = q + \lfloor \log_2 m \rfloor$ and size at most $s_{{2^{n-1}+1}} + \cdots + s_m + 2^{n-1} -1$. Now, the bivariate function $G(x)= (g_1(x),g_2(x))$ can be implemented by a neural network of the form \eqref{eq:def_discriminators} with depth $q+ \lfloor \log_2(m) \rfloor$ and size $s$ such that 
    \begin{align*}
    s \leqslant s_1+\cdots+s_m + 2(2^{n-1} -1) = s_1+\cdots+s_m + 2^n -2 .
    \end{align*}
By concatenating a one neuron layer, we have that the function $f= \max(g_1, g_2)$ can be represented by a neural network of the form \eqref{eq:def_discriminators} with depth $q+\lceil \log_2(m) \rceil$ and size at most $s_1+\cdots+s_m+ 2^n-1 = s_1+\cdots+s_m+ 2^{\lceil \log_2 m \rceil} -1$. The conclusion follows using the inequality $2^{\lceil \log_2 m \rceil} \leqslant 2m$.


\subsection{Proof of Theorem \ref{th:disc_can_represent_piecewise_linear}}
Let $f \in \text{Lip}_1(\mathds R^d)$ that is also $m_f$-piecewise linear. We know that each linear function can be represented by a $1$-neuron neural network verifying Assumption \ref{ass:compactness} (no need for hidden layers). It is easy to see, using a small variant of Proposition \ref{lem:auxiliary_lemma_max_function}, that any collection of $\Tilde{m}$ linear functions with $\Tilde{m} \leqslant m$ can be represented by a neural network of depth $\lceil \log_2(m) \rceil +1$ and size at most $3m-1$. Thus, combining \eqref{eq:piecewise_linear_characterization} with Proposition \ref{lem:auxiliary_lemma_max_function}, for each $k \in \{1, \hdots, M_f \}$ there exists a neural network of the form \eqref{eq:def_discriminators}, verifying Assumption \ref{ass:compactness} and representing the function $\min_{i \in S_k} \ \ell_i$, with depth equal to $\lceil \log_2(m_f) \rceil+1$ (since $|S_k| \leqslant m_f$) and size at most $3m_f-1$.

Using again Proposition \ref{lem:auxiliary_lemma_max_function}, we conclude that there exists a neural network of the form \eqref{eq:def_discriminators}, verifying Assumption \ref{ass:compactness} and representing $f$, with depth$\lceil \log_2(M_f) \rceil+\lceil \log_2(m_f) \rceil+1$ and size at most $3m_f M_f+M_f-1$.

\subsection{Proof of Corollary \ref{cor:piecewise_linear_functions_on_convex_sets}}
According to \citet[][Theorem A.1]{he2018relu}, the function $f$ can be written as 
\begin{equation*}
    f = \underset{1 \leqslant k \leqslant m_f}{\max} \ \underset{i \in S_k}{\min} \ \ell_i,
\end{equation*}
where $|S_k| \leqslant m_f$. Using the same technique of proof as for Theorem \ref{th:disc_can_represent_piecewise_linear}, we find that there exists a neural network of the form \eqref{eq:def_discriminators}, verifying Assumption \ref{ass:compactness} and representing $f$, with depth equal to $2\lceil \log_2(m_f) \rceil+1$ and size at most $3m_f^2+m_f-1$.

\subsection{Proof of Proposition \ref{cor:func_r_to_r}}
Let $f \in \text{Lip}_1(\mathds R)$ that is also $m_f$-piecewise linear. The proof of the first statement is an immediate consequence of Corollary \ref{cor:piecewise_linear_functions_on_convex_sets} since connected subsets of $\mathds{R}$ are also convex. 

As for the second claim of the proposition, considering the case where $f$ is convex, we know from \citet[][Theorem A.1]{he2018relu} that $f$ can be written as
\begin{equation*}
    f = \underset{1 \leqslant k \leqslant m_f}{\max} \ \ell_k.
\end{equation*}
Each function $\ell_k$, $k= 1, \hdots, m_f$, can be represented by a $1$-neuron neural network verifying Assumption \ref{ass:compactness}. Hence, by Proposition \ref{lem:auxiliary_lemma_max_function}, there exists a neural network of the form \eqref{eq:def_discriminators}, verifying Assumption \ref{ass:compactness} and representing $f$, with depth $\lceil \log_2(m_f) \rceil+1$ and size at most $3m_f-1$.

The last claim of the proposition for $m=2^n$ is clear using Proposition \ref{lem:auxiliary_lemma_max_function}.

\subsection{Proof of Lemma  \ref{lem:number_of_pieces}}
The result is proved by induction on $q$. To begin with, in the case $q=2$ we have a neural network with one hidden layer. When applying the GroupSort function with a grouping size $2$, every activation node is defined as the max or min between two different linear functions. The maximum number of breakpoints is equal to the maximum number of intersections, that is $\nu_1/2$. Thus, there is at most $\nu_1/2+1$ pieces.

Now, let us assume that the property is true for a given $q\geqslant3$. Consider a neural network with depth $q$ and widths $\nu_1, \hdots, \nu_{q-1}$. Observe that the input to any node in the last layer is the output of a $\mathds{R} \to \mathds{R}$ GroupSort neural network with depth $(q-1)$ and widths $\nu_1, \hdots, \nu_{q-2}$. Using the induction hypothesis, the input to this node is a function from $\mathds{R} \to \mathds{R}$ with at most $2^{q-3} \times (\nu_1/2+1) \times \cdots \times \nu_{q-2}$ pieces. Thus, after applying the GroupSort function with a grouping size $2$, each node output is a function with at most $2 \times (2^{q-3} \times (\nu_1/2+1) \times \nu_2 \times \cdots \times \nu_{q-2})$. With the final layer, we take an affine combination of $\nu_{q-1}$ functions, each with at most $2^{q-2} \times (\nu_1/2+1) \times \nu_2 \times \cdots \times \nu_{q-2}$ pieces. In all, we therefore get at most $2^{q-2} \times (\nu_1/2+1) \times \nu_2 \times \cdots \times \nu_{q-1}$ pieces. The induction step is completed.  

\subsection{Proof of Corollary \ref{cor:lower_bound_for_nn}}
Let $f$ be an $m_f$-piecewise linear function. For a neural network of depth $q$ and widths $\nu_1, \hdots, \nu_q$ representing $f$, we have, by Lemma \ref{lem:number_of_pieces},
\begin{equation*}
    2^{q-1} \times (\nu_1/2+1) \times \cdots \times \nu_{q-1} \geqslant m_f.
\end{equation*}
By the inequality of arithmetic and geometric means, minimizing the size $s=\nu_1/2+ \cdots+\nu_k$ subject to this constraint, means setting $\nu_1/2+1=\nu_2=\cdots=\nu_k$. This implies that $s \geqslant \frac{1}{2} (q-1) m_f^{1/(q-1)}$.

\subsection{Proof of Theorem \ref{th:approximating_any_lipschitz_function}}
The proof follows the one from \citet[][Theorem 3]{cooper1995learning}. Tesselate $[0,1]^d$ by cubes of side $s= \varepsilon/(2\sqrt{d})$ and denote by $n= (\lceil 1/s \rceil)^d$ the number of cubes in the tesselation. Choose $n$ data points, one in each different cube. Then any Delaunay sphere will have a radius $R < \varepsilon/2M_f$. Now, construct $\Tilde{f}$ by linearly interpolating between values of $f$ over the Delaunay simplices. According to \citet{seidel1995upper}, the number $m_f$ of subdomains is $O(n^{d/2})$ and each of them is convex. Besides, by \citet[][Lemma 2]{cooper1995learning}, $\Tilde{f}$ guarantees an approximation error $\|f- \Tilde{f}\|_\infty \leqslant \varepsilon$. 
    
Using Corollary \ref{cor:piecewise_linear_functions_on_convex_sets}, we know that there exists a neural network of the form \eqref{eq:def_discriminators} verifying Assumption \ref{ass:compactness} and representing $\Tilde{f}$. Besides, its depth is $2 \lceil \log_2(m_f) \rceil+1$ and its size is at most $3m_f^2+m_f-1$. Consequently, we have that the depth of the neural network is $2\lceil \log_2(m_f) \rceil +1= O(d^2 \log_2(\frac{2\sqrt{d}}{\varepsilon}))$ and the size at most $O(m^{2}) = O((\frac{2\sqrt{d}}{\varepsilon})^{d^2})$.

\subsection{Proof of Proposition \ref{cor:approximating_real_valued_lipschitz_function}}
Let $f \in \text{Lip}_1([0,1])$ and $f_m$ be the piecewise linear interpolation of $f$ with the following $2^m+1$ breakpoints: $k/2^m$, $k= 0,\hdots,2^m$. We know that the function $f_m$ approximates $f$ with an error $\varepsilon_m \leqslant 2^{-m}$. In particular, for any $m \geqslant \log_2(1/\varepsilon)$, we have $\varepsilon_m \leqslant \varepsilon$. Besides, for any $m$, $f_m$ is a $1$-Lipschitz function defined on $[0,1]$, piecewise linear on $2^m$ subdomains. Thus, according to Proposition \ref{cor:func_r_to_r}, there exists a neural network of the form \eqref{eq:def_discriminators}, verifying Assumption \ref{ass:compactness} and representing $f_m$, with depth $2m+1$ and size at most $3 \times 2^{2m}+2^m-1$. Taking $m=\lceil \log_2 (1/\varepsilon) \rceil$ shows the desired result.

Let $\varepsilon>0$, let $f$ be a convex (or concave) function in $\text{Lip}_1([0,1])$, and let $f_m$ be the piecewise linear interpolation of $f$ with the following $2^m+1$ breakpoints: $k/2^m$, $k= 0,\hdots,2^m$. The function $f_m$ approximates $f$ with an error $\varepsilon_m= 2^{-m}$. In particular, for any $m \geqslant \log_2(1/\varepsilon)$, we have $\varepsilon_m \leqslant \varepsilon$. Besides, for any $m$, $f_m$ is a $2^m$-piecewise linear convex function defined on $[0,1]$. Hence, by Proposition \ref{cor:func_r_to_r}, there exists a neural network of the form \eqref{eq:def_discriminators}, verifying Assumption \ref{ass:compactness} and representing $f_m$, with depth $m+1$ and size at most $2 \times 2^m-1$. Taking  $m=\lceil \log_2(1/\varepsilon) \rceil$ leads to the desired result. 

\subsection{Proof of Proposition \ref{lem:auxiliary_lemma_max_function_GSk}}
We prove the result by using an induction on $n$. The case where $n=1$ and thus $m=k^1$ is true since the function $f = \max(f_1, \hdots, f_k)$ can be represented by a neural network of the form \eqref{eq:def_discriminators} with grouping size $k$, depth $q+1$, and size $s_1+\cdots+s_k+1$. Now, let $m=k^n$ with $n > 1$. We have that $\lfloor m/k \rfloor = \lceil m/k \rceil = m/k = k^{n-1}$. Let 
$g_1 = \max(f_1, \hdots, f_{m/k}), g_2= \max(f_{m/k+1}, \hdots, f_{2m/k}), \hdots, g_k=\max(f_{((k-1)m/k)+1}, \hdots, f_{m})$. By the induction hypothesis, $g_1, \hdots, g_k$ can all be represented by neural networks of the form \eqref{eq:def_discriminators} with grouping size $k$, width depths equal to $q+n-1$ and sizes at most $s_1 + \cdots + s_{m/k} + \frac{k^{n-1}-1}{k-1}, \hdots, s_{(k-1)m/k+ 1} + \cdots + s_{m} + \frac{k^{n-1}-1}{k-1}$, respectively.

Consequently, the function $G(x)= (g_1(x), \hdots, g_k(x))$ can be implemented by a neural network of the form \eqref{eq:def_discriminators} with grouping size $k$, depth $q+ n-1$, and size at most $s_1+\cdots+s_m+m-2$. Finally, by concatenating a one neuron layer, we see that the function $f= \max(g_1, \hdots, g_k)$ can be represented by a neural network of the form \eqref{eq:def_discriminators} with depth $q+n=q+\log_k(m)$ and size at most
\begin{equation*}
    s_1+\cdots+s_m+k\Big(\frac{k^{n-1}-1}{k-1}\Big)+1 = s_1+\cdots+s_m+ \frac{k^n-1}{k-1}= s_1+\cdots+s_m+\frac{m-1}{k-1}.
\end{equation*}

\subsection{Proof of Corollary \ref{cor:piecewise_linear_functions_on_convex_sets_arbitrary_k}} 
According to \citet[][Theorem A.1]{he2018relu}, the function $f$ can be written as 
\begin{equation*}
    f = \underset{1 \leqslant k \leqslant m_f}{\max} \ \underset{i \in S_k}{\min} \ \ell_i,
\end{equation*}
where $|S_k| \leqslant m_f$ and $m_f=k^n$ for some $n\geqslant1$. 
It is easy to see, using a small variant of Proposition \ref{lem:auxiliary_lemma_max_function_GSk}, that any collection of $\Tilde{m}$ linear functions with $\Tilde{m} \leqslant m_f$ can be represented by a neural network of depth $\log_k(m) +1$ and size at most $\frac{m_f-1}{k-1}$.
Therefore, by Proposition \ref{lem:auxiliary_lemma_max_function_GSk}, there exists a neural network verifying Assumption \ref{ass:compactness} with grouping size $k$ representing $\underset{i \in S_k}{\min} \ \ell_i$ with depth $\log_k(m)+1$ and size at most $\frac{m_f-1}{k-1}$.

Using again Proposition \ref{lem:auxiliary_lemma_max_function_GSk}, we find that there exists a neural network, verifying Assumption \ref{ass:compactness}, with grouping size $k$, representing $f$ with depth $2\log_k(m_f)+1$ and size at most
\begin{equation*}
    m_f\Big(\frac{m_f-1}{k-1}\Big)+\frac{m_f-1}{k-1} = \frac{m_f^2-1}{k-1}.
\end{equation*}

\subsection{Proof of Lemma \ref{lem:number_of_pieces__GSk}}
The result is proved by induction on $q$. To begin with, in the case $q=2$ we have a neural network with one hidden layer. When applying the GroupSort function with a grouping size $k$, the maximum number of breakpoints is equal to the maximum number of intersections of linear functions. In each group of $k$ functions, there are at most $\frac{k(k-1)}{2}$ intersections. Thus, there are at most $\frac{k(k-1)}{2} \times \frac{\nu_1}{k} = \frac{(k-1)\nu_1}{2}$ breakpoints, that is $\frac{(k-1)\nu_1}{2}+1$ pieces.

Now, let us assume that the property is true for a given $q\geqslant3$. Consider a neural network with depth $q$ and widths $\nu_1, \hdots, \nu_{q-1}$. Observe that the input to any node in the last layer is the output of a $\mathds{R} \to \mathds{R}$ GroupSort neural network with depth $(q-1)$ and widths $\nu_1, \hdots, \nu_{q-2}$. Using the induction hypothesis, the input to this node is a function from $\mathds{R} \to \mathds{R}$ with at most $k^{q-3} \times (\frac{(k-1)\nu_1}{2}+1) \times \cdots \times \nu_{q-2}$ pieces. Thus, after applying the GroupSort function with a grouping size $k$, each node output is a function with at most $k \times (k^{q-3} \times (\frac{(k-1)\nu_1}{2}+1) \times \nu_2 \times \cdots \times \nu_{q-2})$. With the final layer, we take an affine combination of $\nu_{q-1}$ functions, each with at most $k^{q-2} \times (\frac{(k-1)\nu_1}{2}+1) \times \nu_2 \times \cdots \times \nu_{q-2}$ pieces. In all, we therefore get at most $k^{q-2} \times (\frac{(k-1)\nu_1}{2}+1) \times \nu_2 \times \cdots \times \nu_{q-1}$ pieces. The induction step is completed.

\subsection{Proof of Theorem \ref{th:approximating_any_lipschitz_function_arbitrary_k}}
The proof of Theorem \ref{th:approximating_any_lipschitz_function_arbitrary_k} is straightforward and follows the one of Theorem \ref{th:approximating_any_lipschitz_function}
combined with the result obtained in Corollary \ref{cor:piecewise_linear_functions_on_convex_sets_arbitrary_k}.

\subsection{Proof of Proposition \ref{cor:approximating_real_valued_lipschitz_function_arbitrary_k}}
Let $f \in \text{Lip}_1([0,1])$ and $f_m$ be the piecewise linear interpolation of $f$ with the following $k^n+1$ breakpoints: $i/k^n$, $k= 0,\hdots,k^n$. We know that the function $f_m$ approximates $f$ with an error $\varepsilon_m \leqslant k^{-n}$. In particular, for any $n \geqslant \log_k(1/\varepsilon)$, we have $\varepsilon_n \leqslant \varepsilon$. Besides, for any $n$, $f_{k^n}$ is a $1$-Lipschitz function defined on $[0,1]$, piecewise linear on $k^n$ subdomains. Thus, according to Corollary \ref{cor:piecewise_linear_functions_on_convex_sets_arbitrary_k}, there exists a neural network of the form \eqref{eq:def_discriminators}, verifying Assumption \ref{ass:compactness} and representing $f_{k^n}$, with grouping size $k$, depth $2n+1$, and size at most $\frac{k^{2n}-1}{k-1}$. Taking $n=\lceil \log_k(1/\varepsilon) \rceil$ shows the desired result. 

\section{Experiments: Extended comparison between GroupSort and ReLU networks}
We provide in this section further results and details on the experiments ran in Section \ref{section:experiments}. 

\subsection{Task 1: Approximating functions}

\begin{figure}[h]
    \centering
    {
        \includegraphics[width=0.31\linewidth]{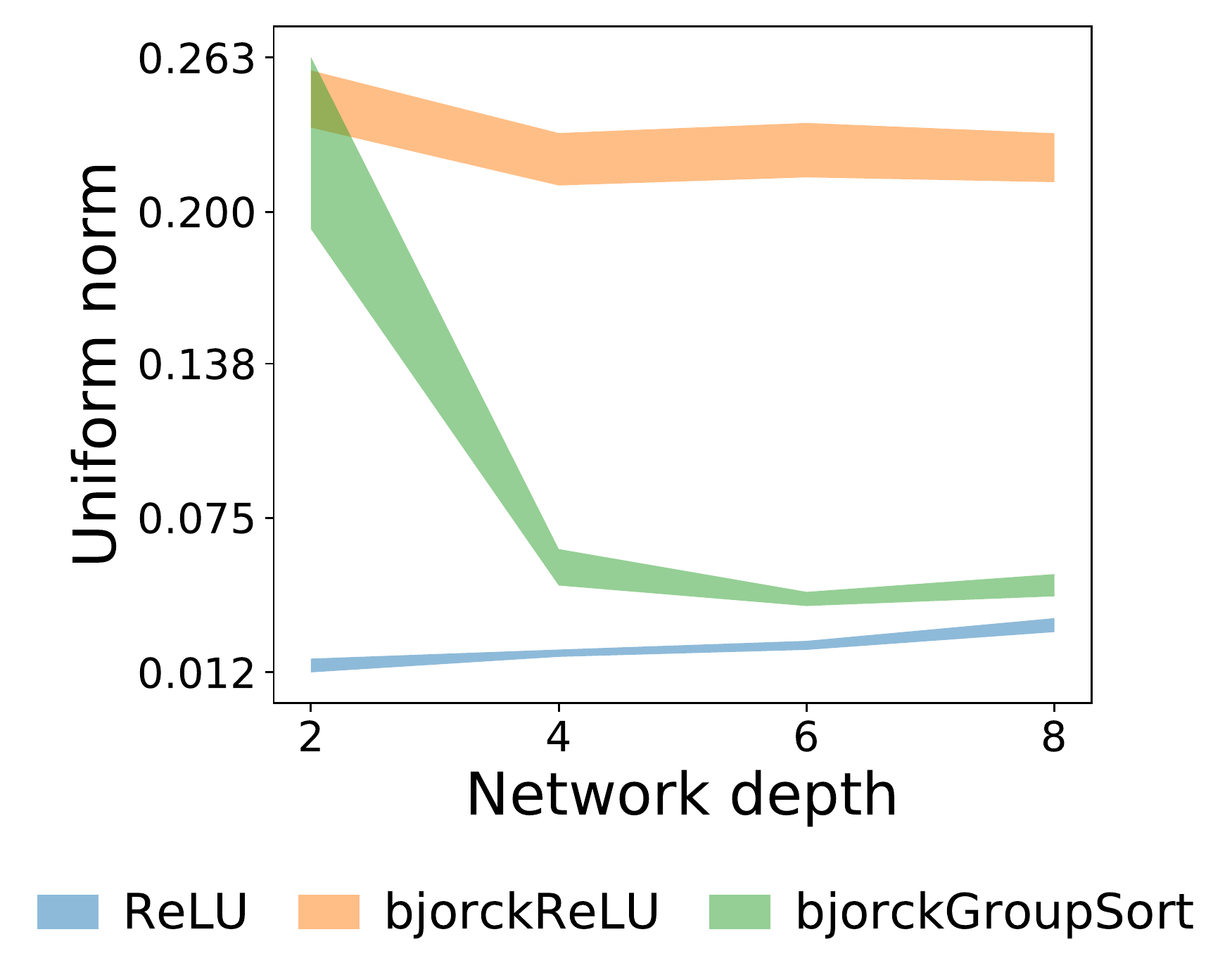}
    }
    {
        \includegraphics[width=0.31\linewidth]{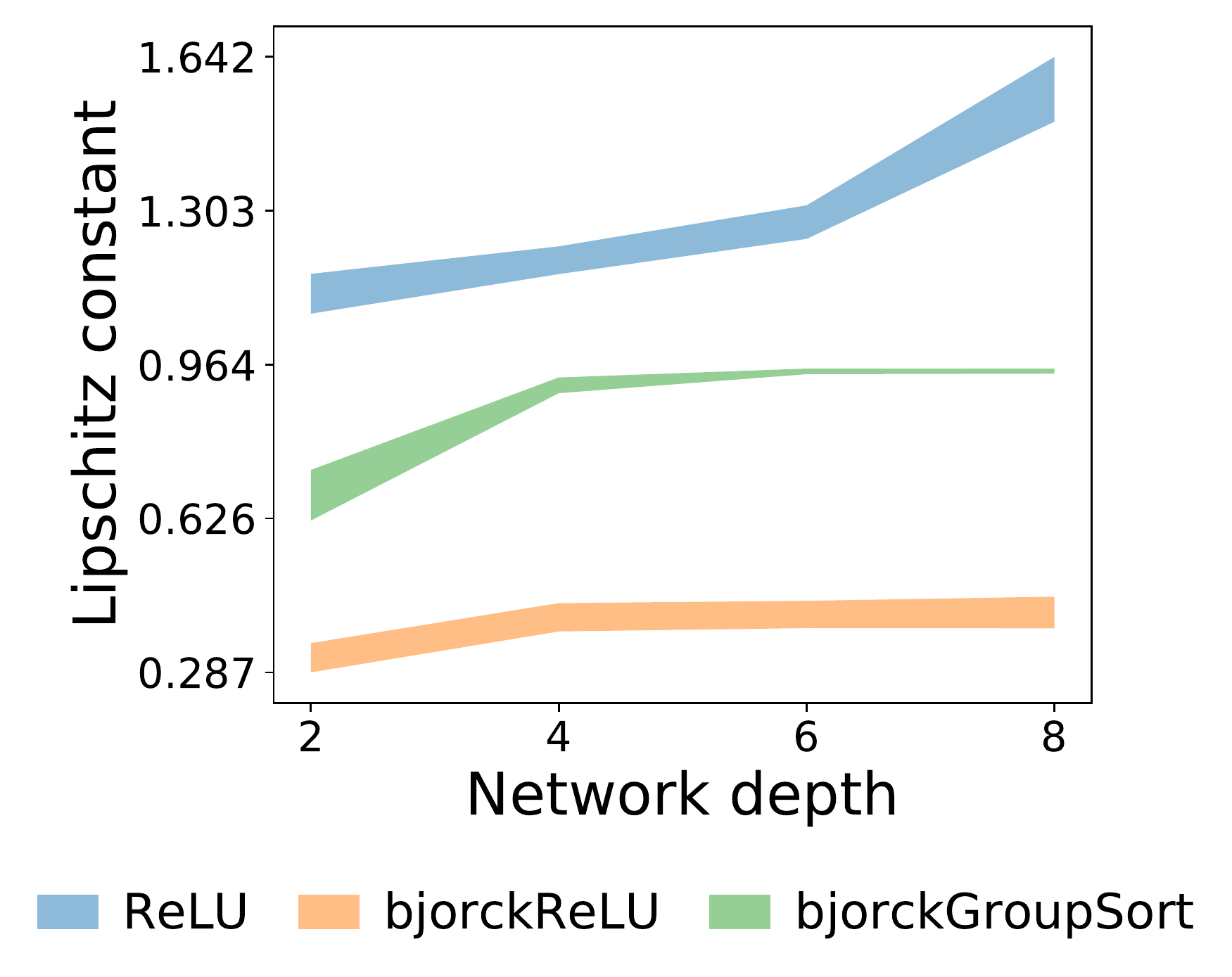}
    }
    {
        \includegraphics[width=0.31\linewidth]{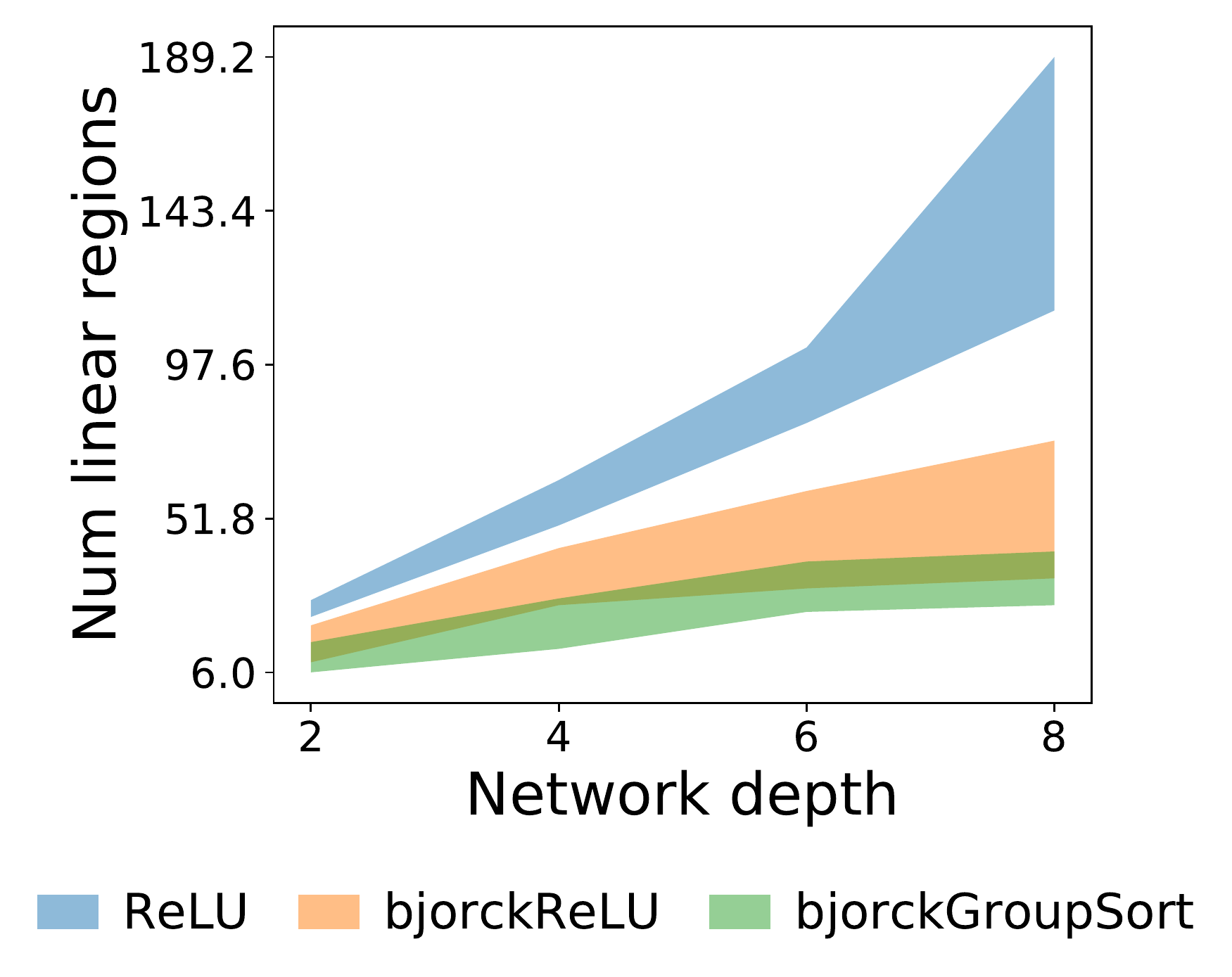}
    }
    \caption{Estimating the 6-piecewise linear function in the model $Y=f(X)$, with a dataset of size $n=100$ (the thickness of the line represents a $95\%$-confidence interval).}
    \label{fig:6pwl_no_noise}
\end{figure}

\begin{figure}[h]
    \centering
    \subfloat[ReLU]
    {
        \includegraphics[width=0.31\linewidth]{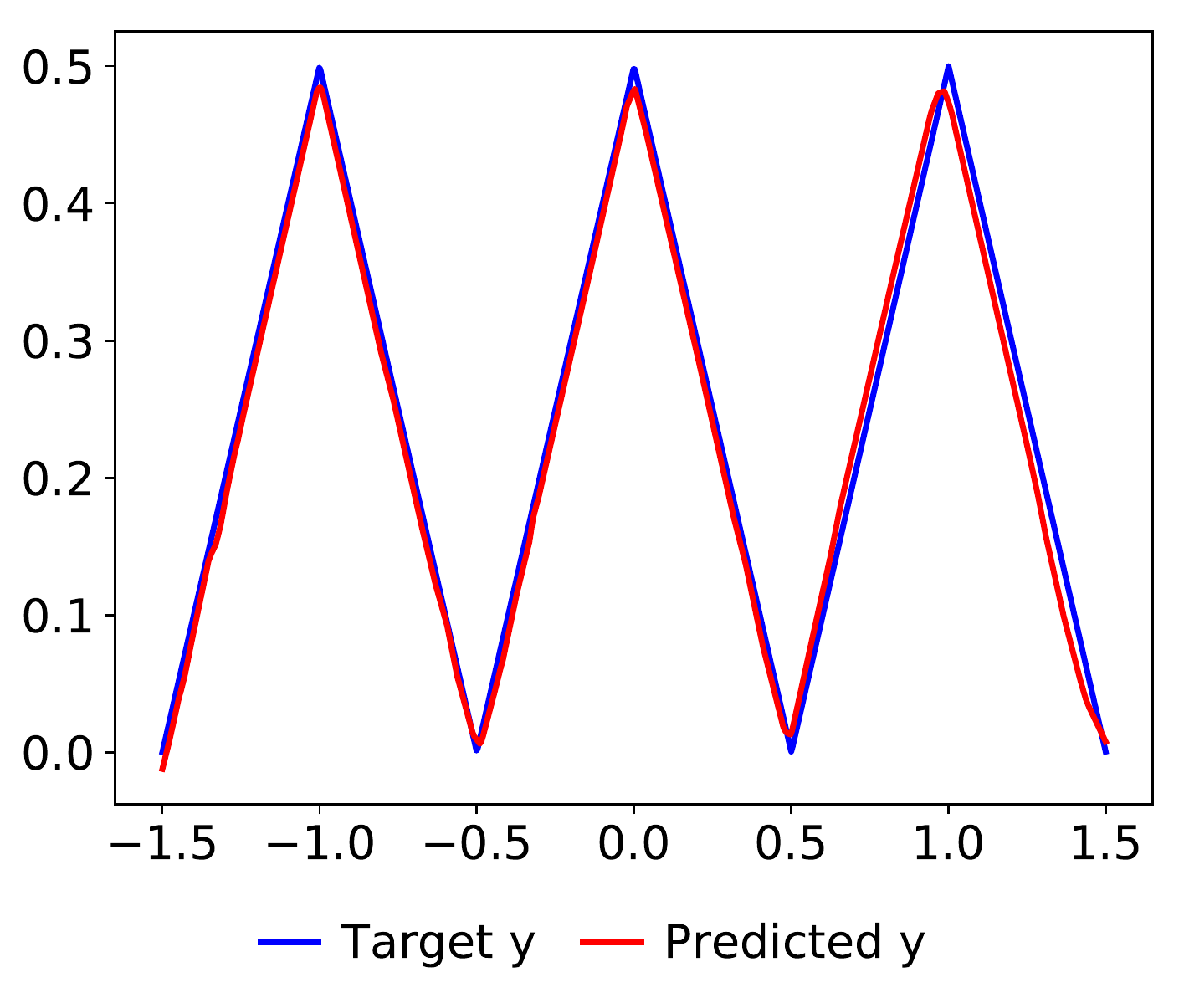}
    }
    \subfloat[bjorckReLU]
    {
        \includegraphics[width=0.31\linewidth]{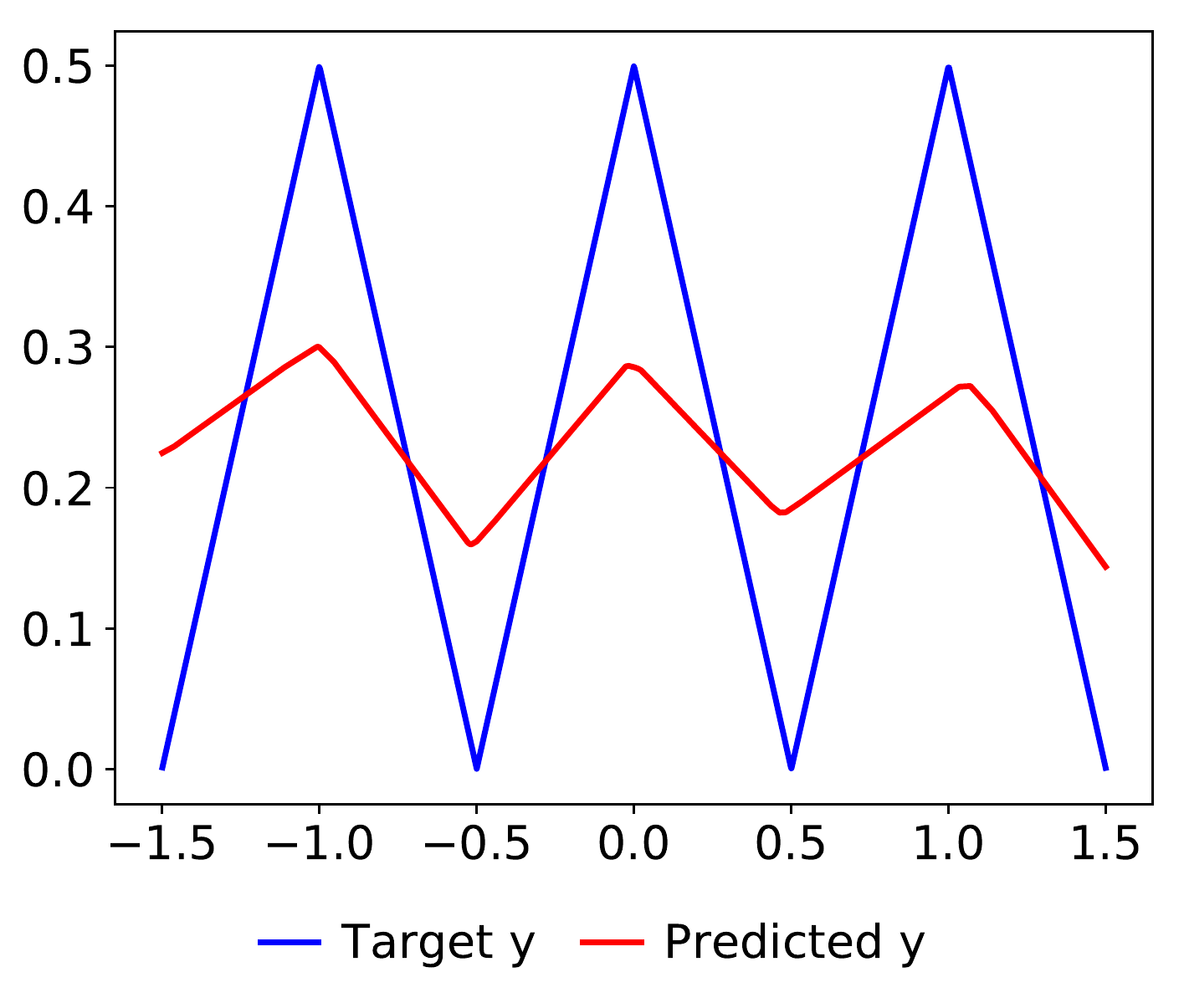}
    }
    \subfloat[bjorckGroupSort]
    {
        \includegraphics[width=0.31\linewidth]{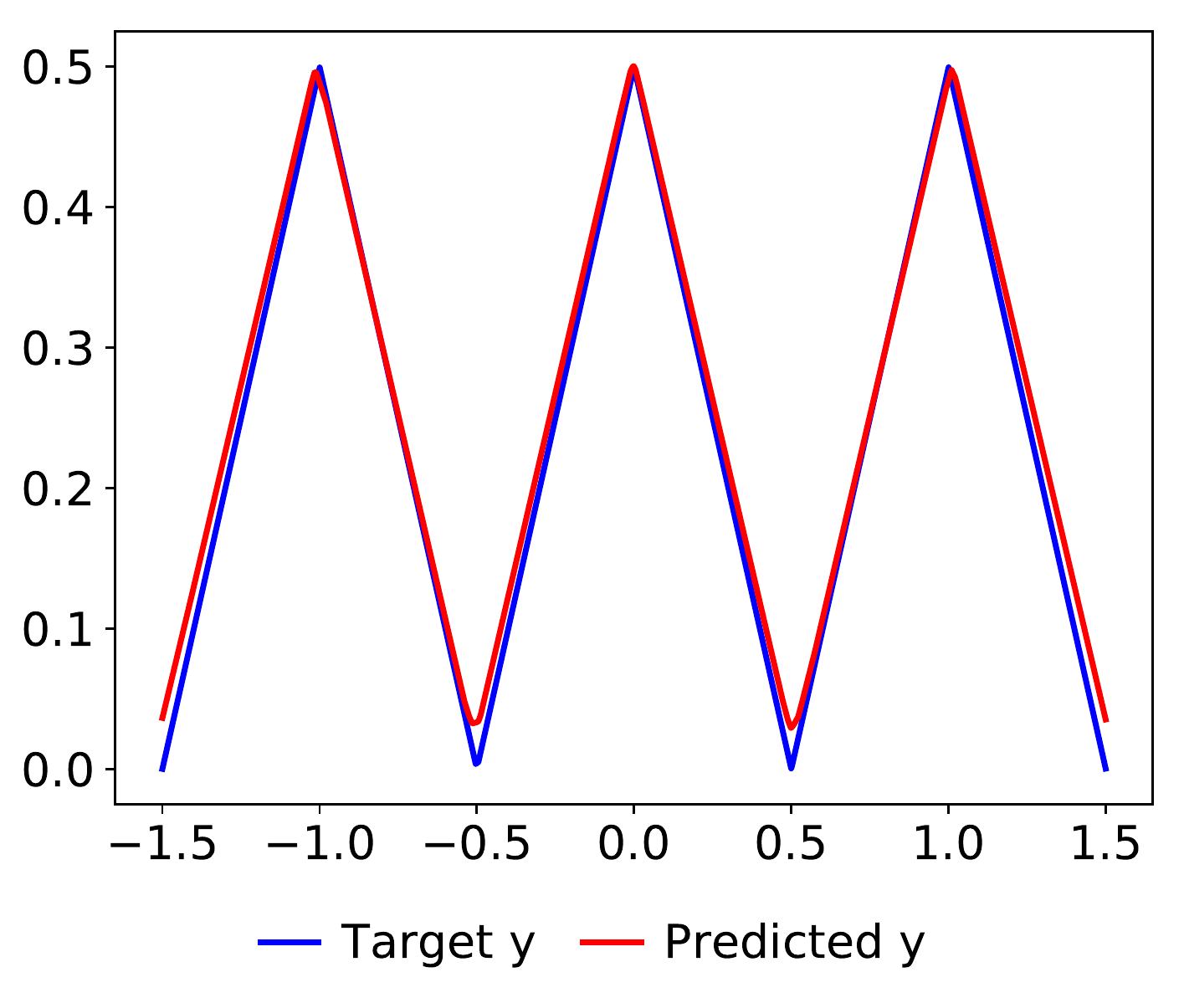}
    }
    \caption{Reconstructing the 6-piecewise linear function in the model $Y=f(X)$, with a dataset of size $n=100$.}
    \label{fig:6pwl_no_noise_examples}
\end{figure}

\begin{figure}[H]
    \centering
    {
        \includegraphics[width=0.31\linewidth]{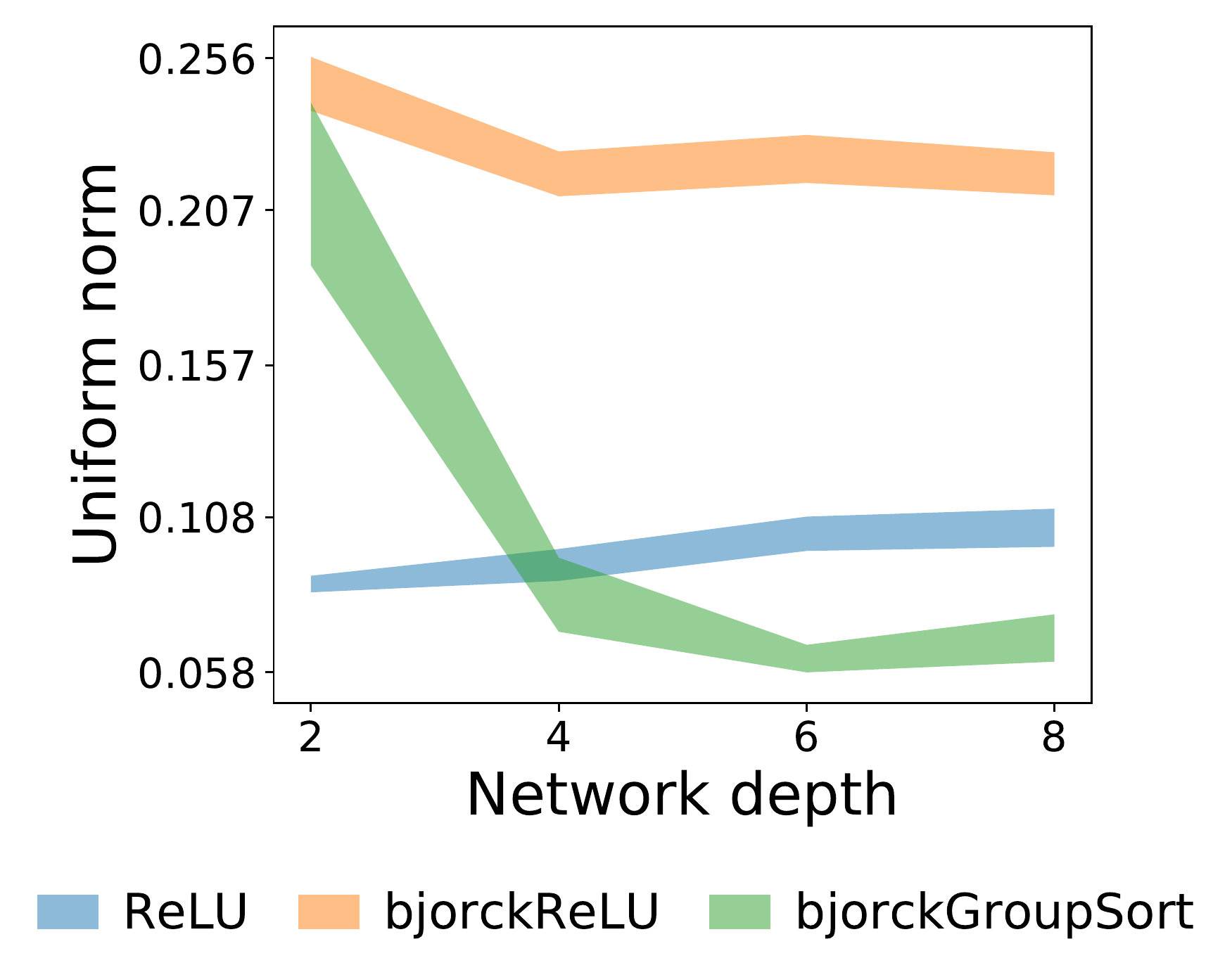}
    }
    {
        \includegraphics[width=0.31\linewidth]{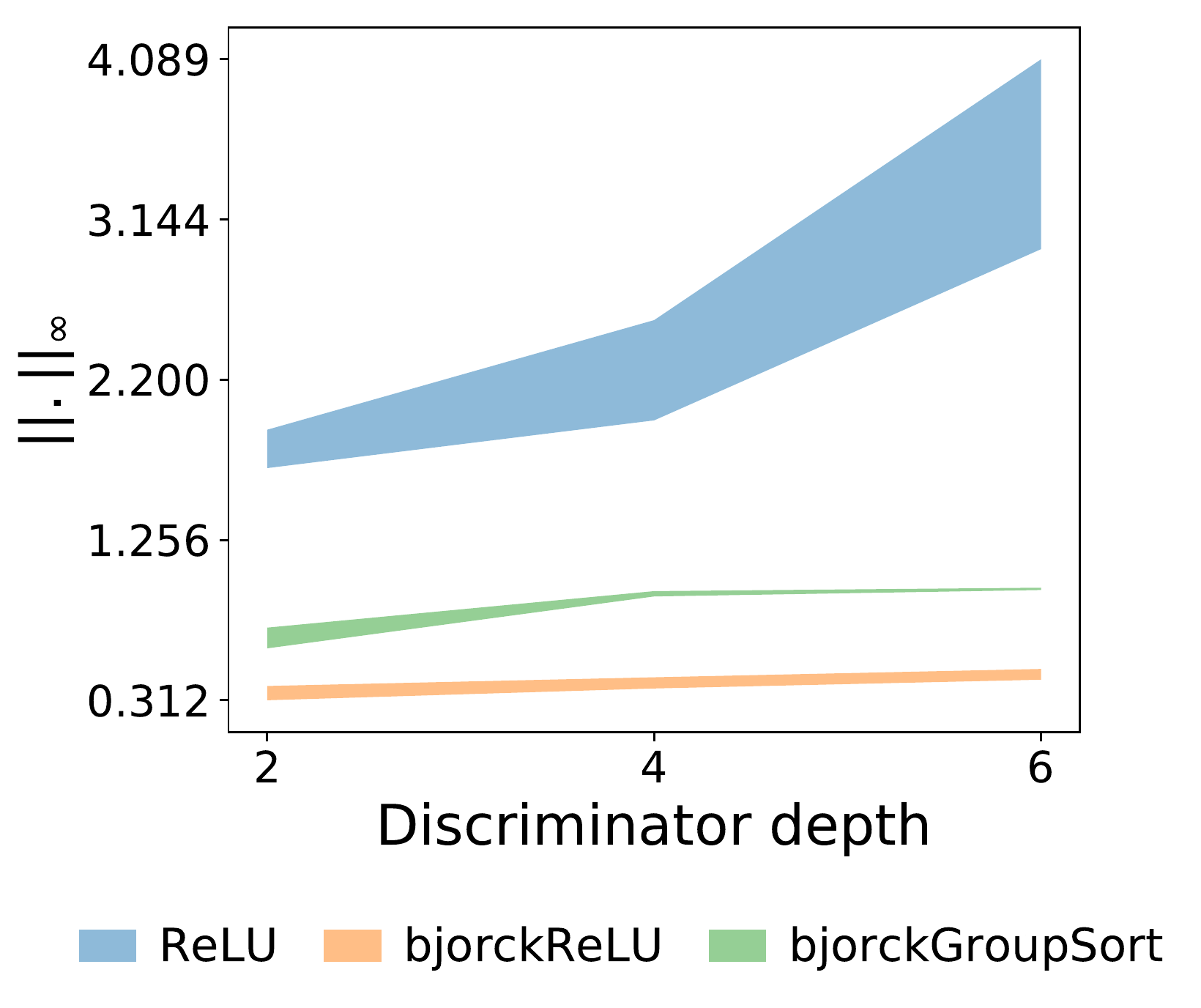}
    }
    {
        \includegraphics[width=0.31\linewidth]{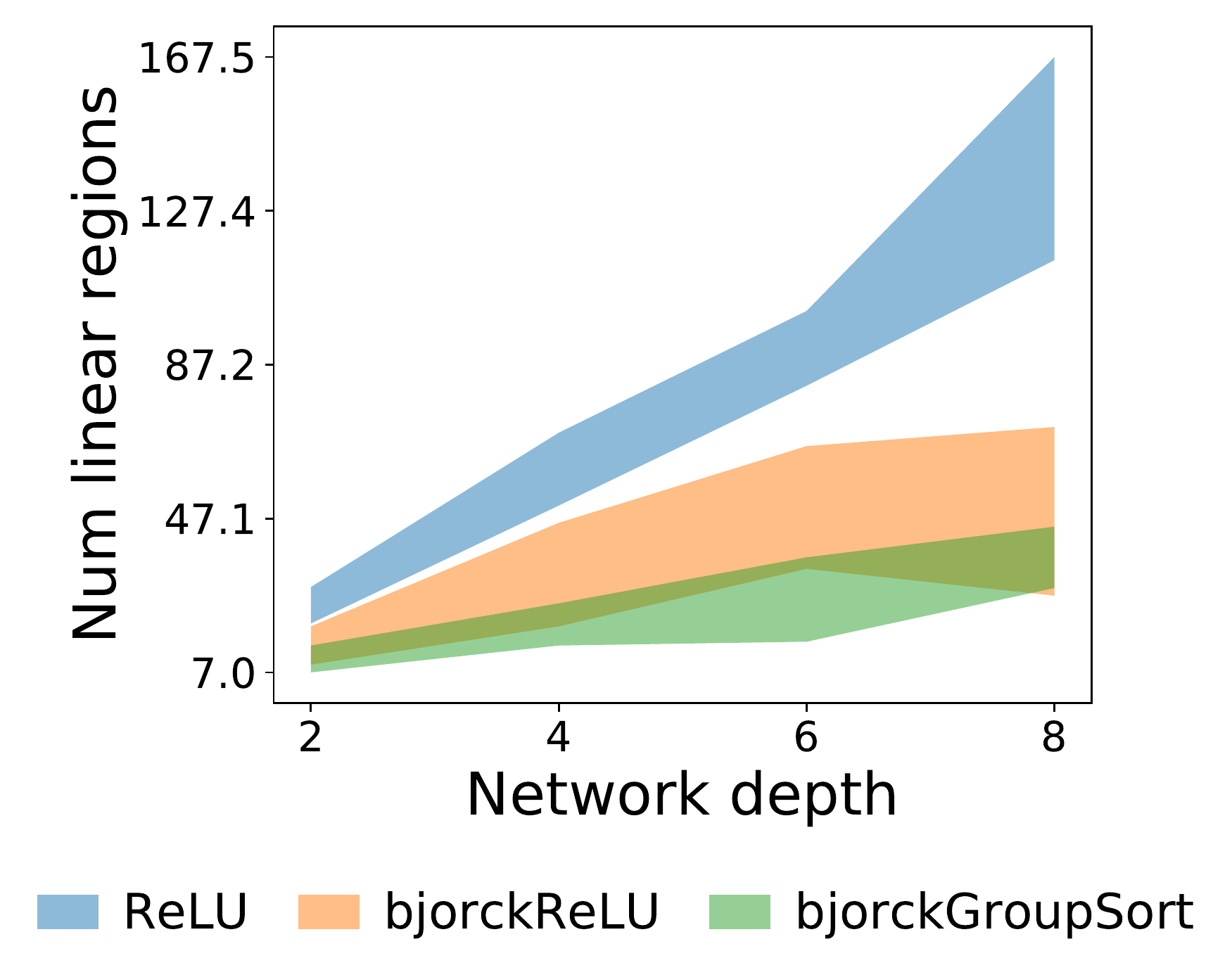}
    }
    \caption{Estimating the 6-piecewise linear function in the model $Y=f(X)+\varepsilon$, with a dataset of size $n=100$ (the thickness of the line represents a $95\%$-confidence interval).}
    \label{fig:6pwl_noise}
\end{figure}

\begin{figure}[H]
    \centering
    \subfloat[ReLU]
    {
        \includegraphics[width=0.31\linewidth]{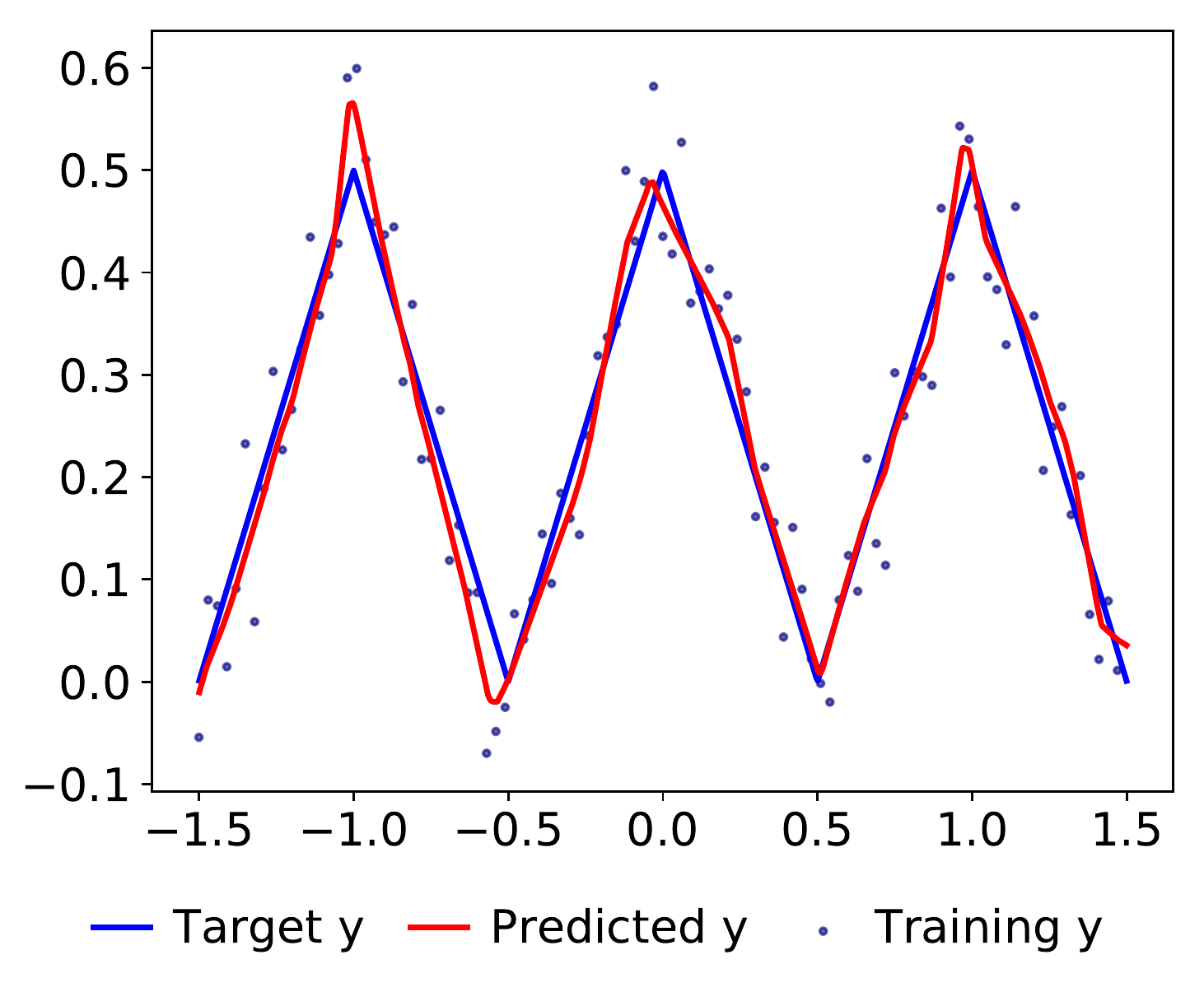}
    }
    \subfloat[bjorckReLU]
    {
        \includegraphics[width=0.31\linewidth]{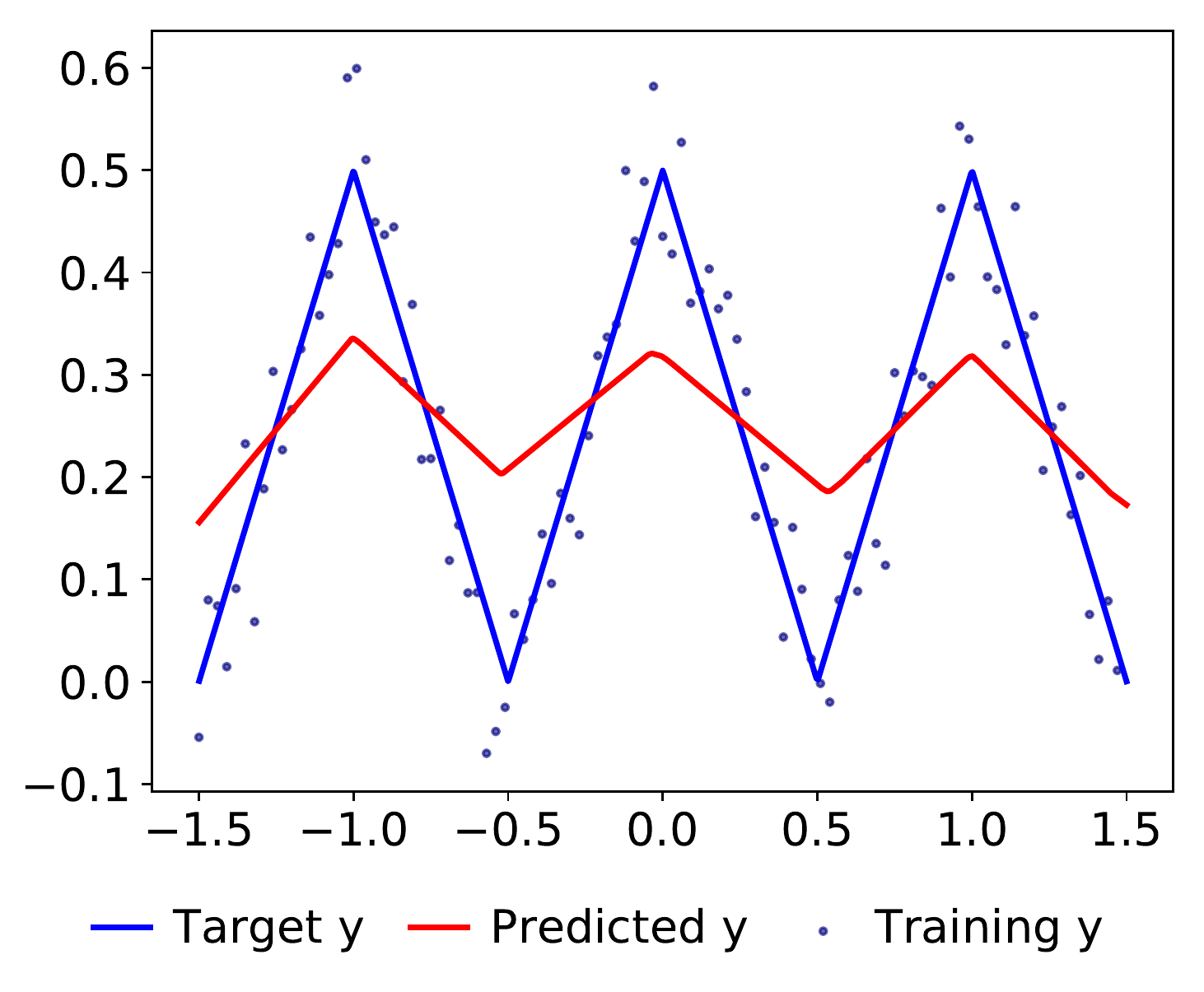}
    }
    \subfloat[bjorckGroupSort]
    {
        \includegraphics[width=0.31\linewidth]{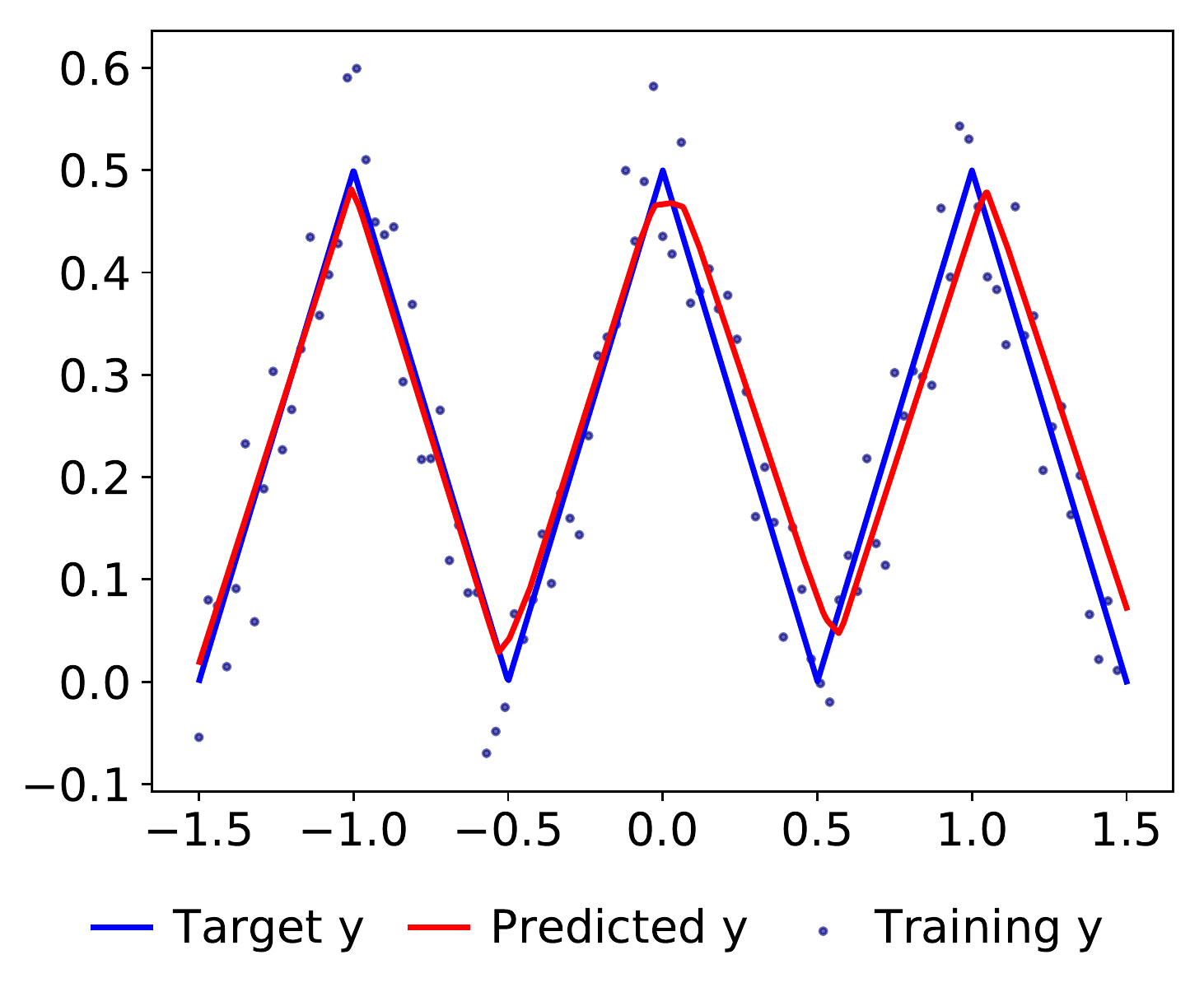}
    }
    \caption{Reconstructing the 6-piecewise linear function in the model $Y=f(X)+\varepsilon$, with a dataset of size $n=100$.}
    \label{fig:6pwl_noise_examples}
\end{figure}

\paragraph{Piecewise linear functions.} We complete the experiments of Section \ref{section:experiments} by estimating the $6$-piecewise linear function $f$ in the model $Y=f(X)$ (noiseless case, see Figure \ref{fig:6pwl_no_noise} and Figure \ref{fig:6pwl_no_noise_examples}) and in the model $Y=f(X)+\varepsilon$ (noisy case, see Figure \ref{fig:6pwl_noise} and Figure \ref{fig:6pwl_noise_examples}). Recall that in both cases, $X$ follows a uniform distribution on $[-1.5, 1.5]$ and the sample size is $n=100$.

\begin{figure}[H]
    \centering
    \subfloat[ReLU]
    {
        \includegraphics[width=0.32\linewidth]{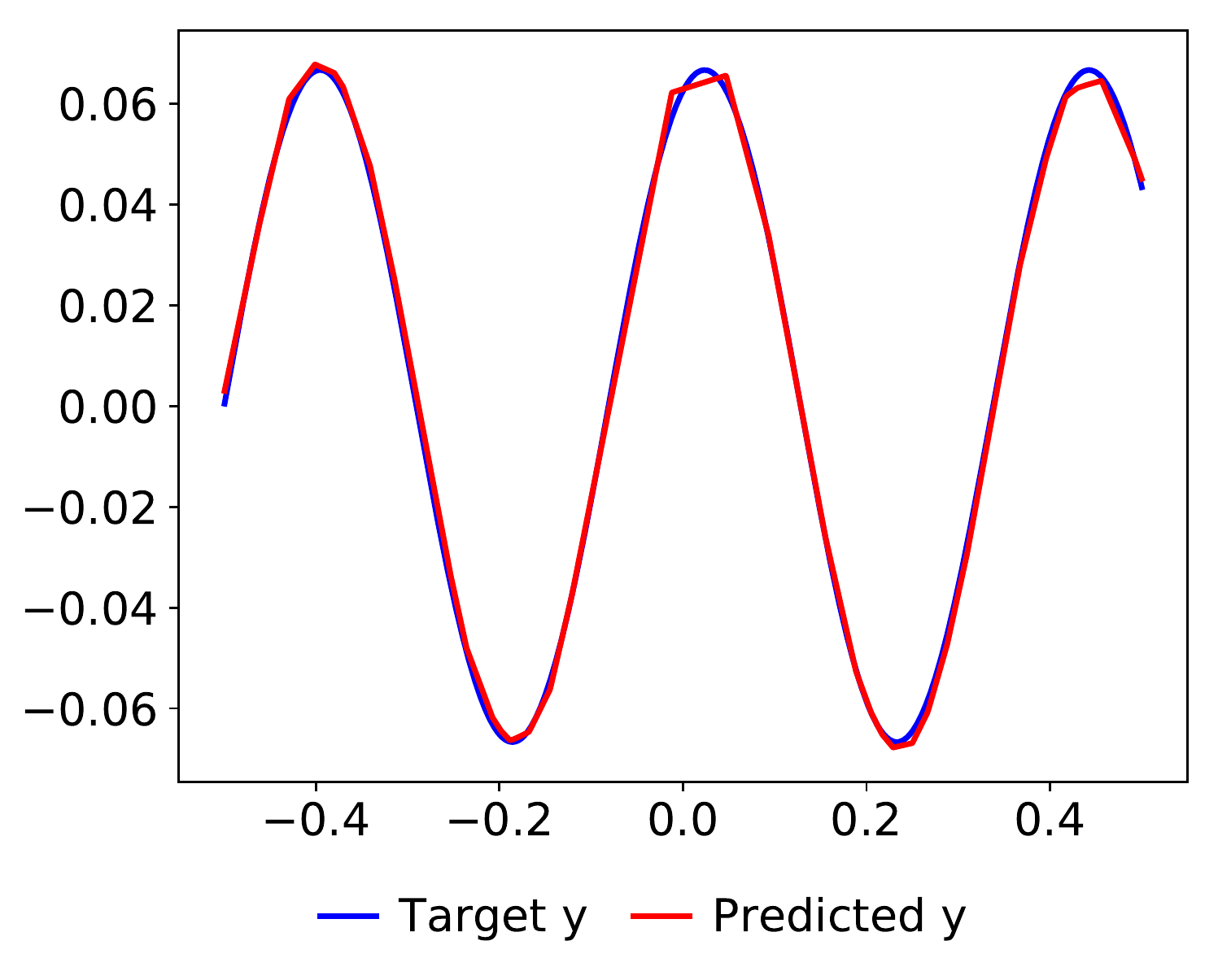}
    }
    \subfloat[bjorckReLU]
    {
        \includegraphics[width=0.32\linewidth]{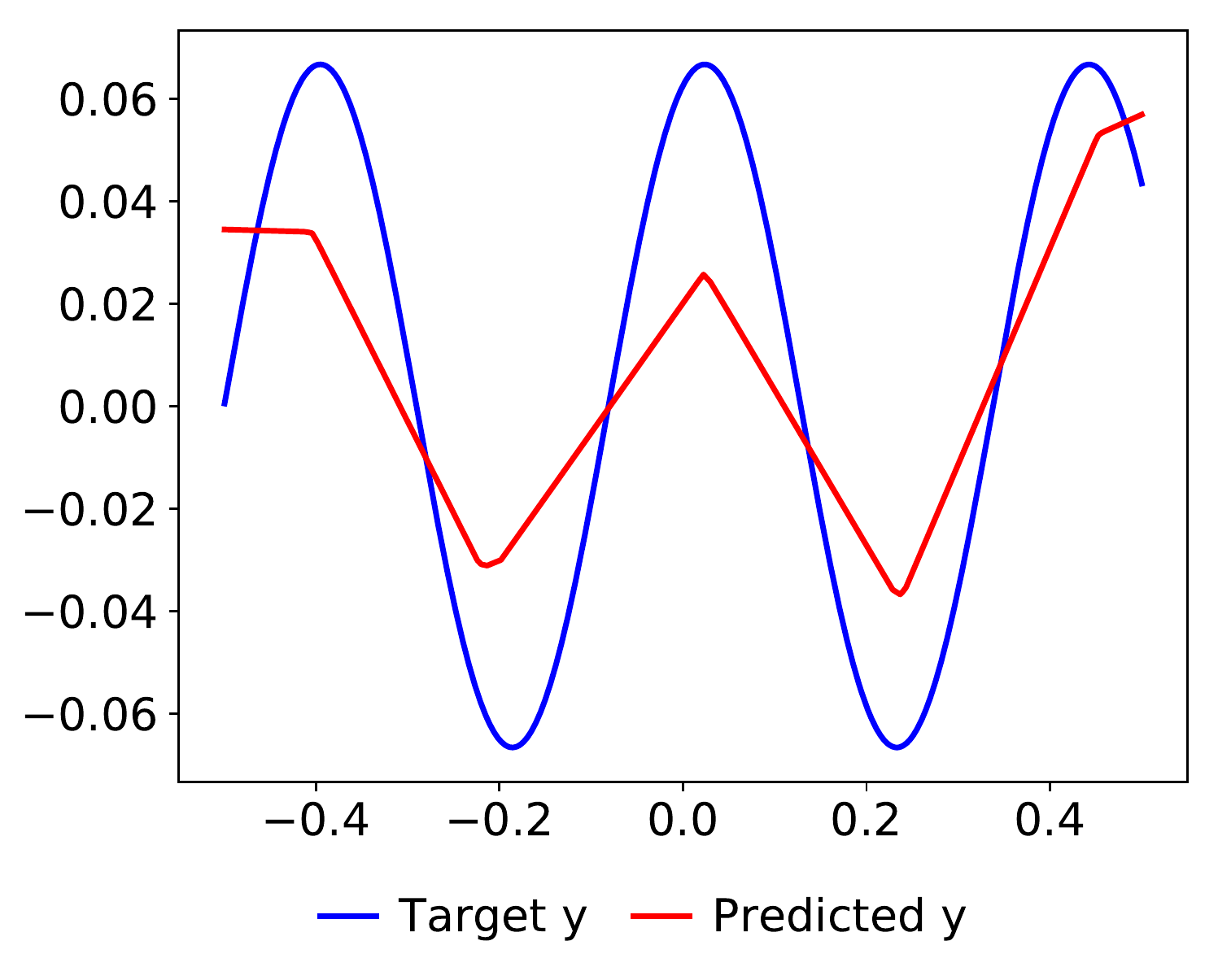}
    }
    \subfloat[bjorckGroupSort]
    {
        \includegraphics[width=0.32\linewidth]{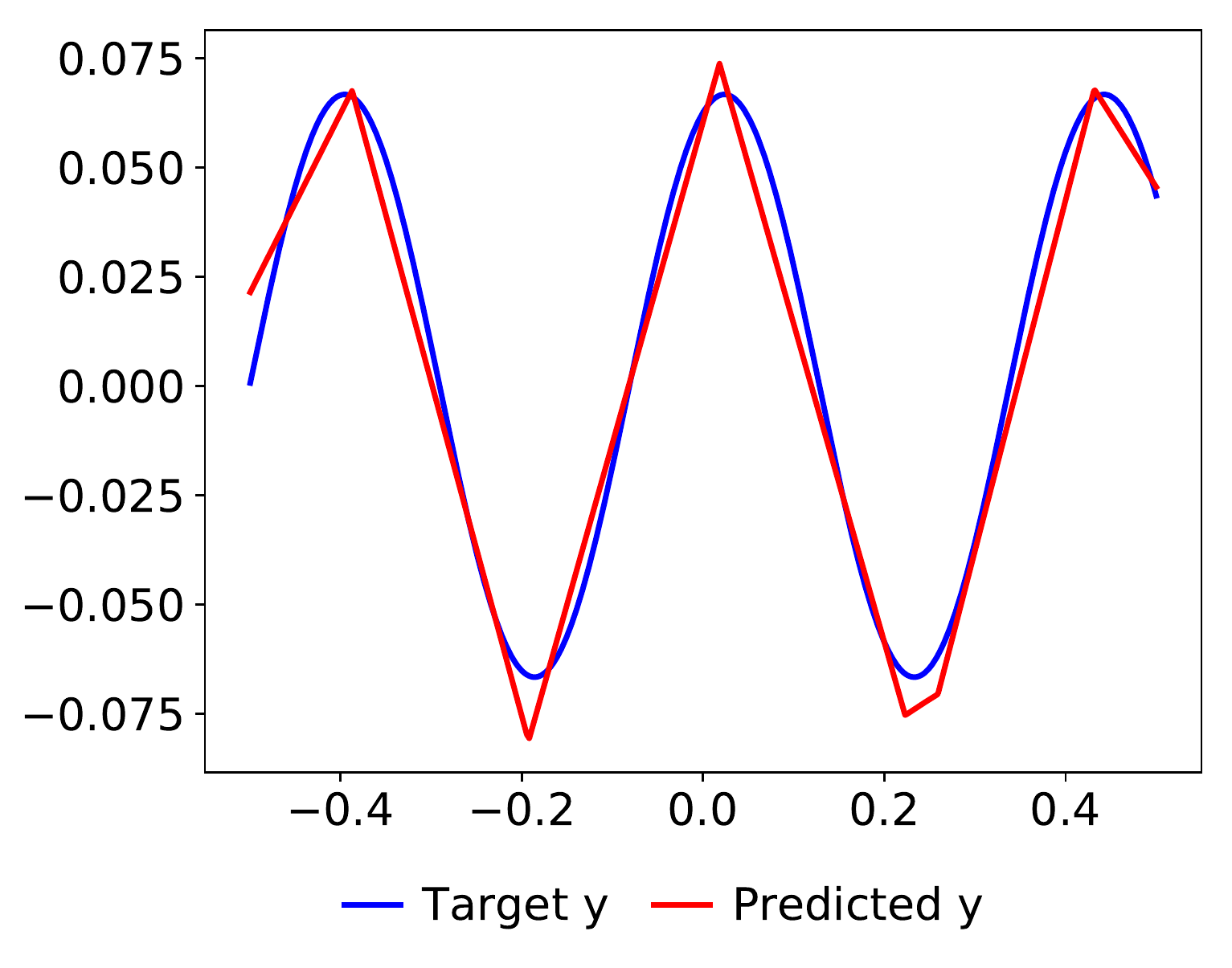}
    }
    \caption{Reconstructing the function $f(x)=(1/15)\sin(15x)$ in the model $Y=f(X)$, with a dataset of size $n=100$.}
    \label{fig:approx_sinus_without_noise}
\end{figure}

\begin{figure}[H]
    \centering
    \subfloat[ReLU]
    {
        \includegraphics[width=0.32\linewidth]{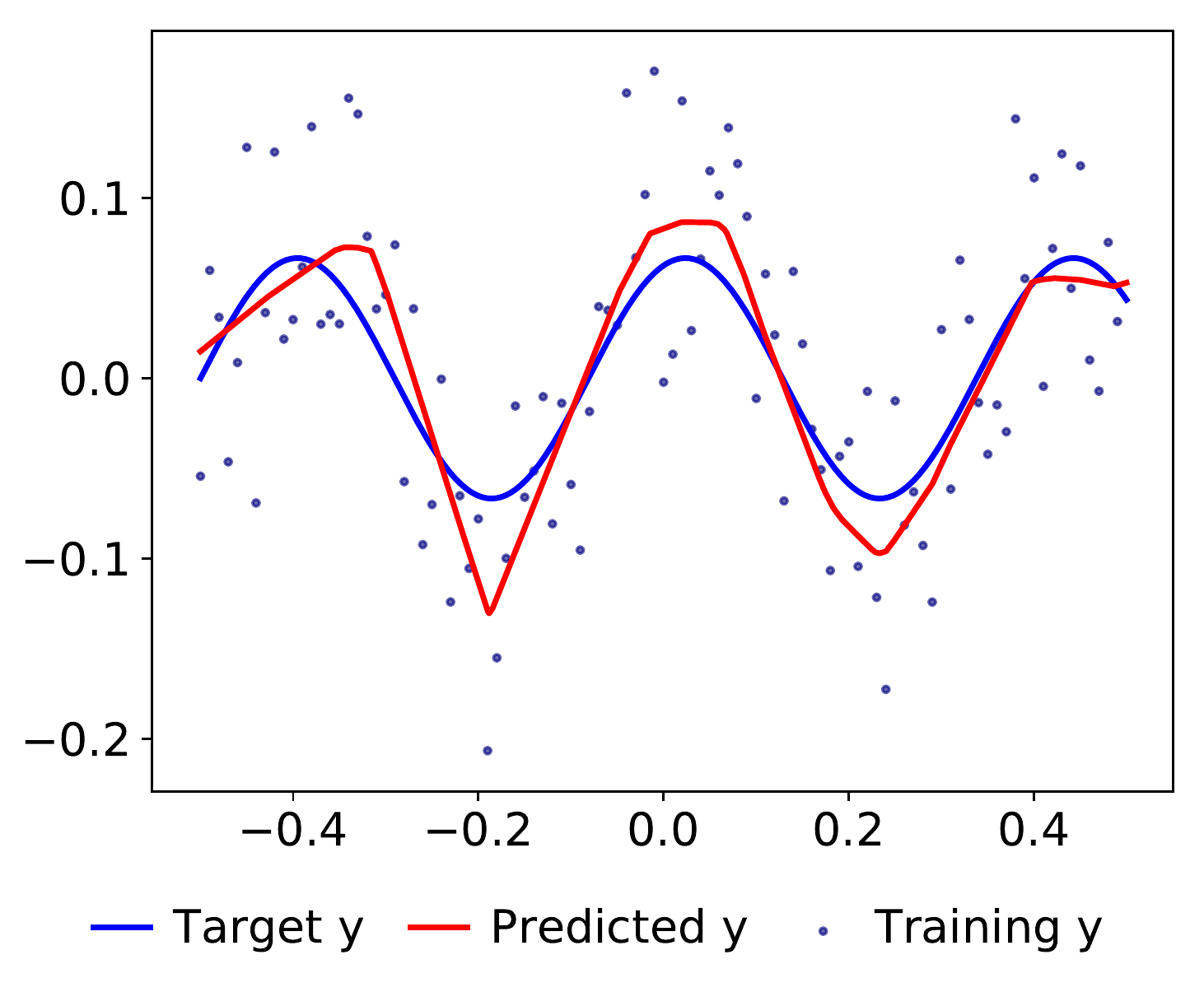}
    }
    \subfloat[bjorckReLU]
    {
        \includegraphics[width=0.32\linewidth]{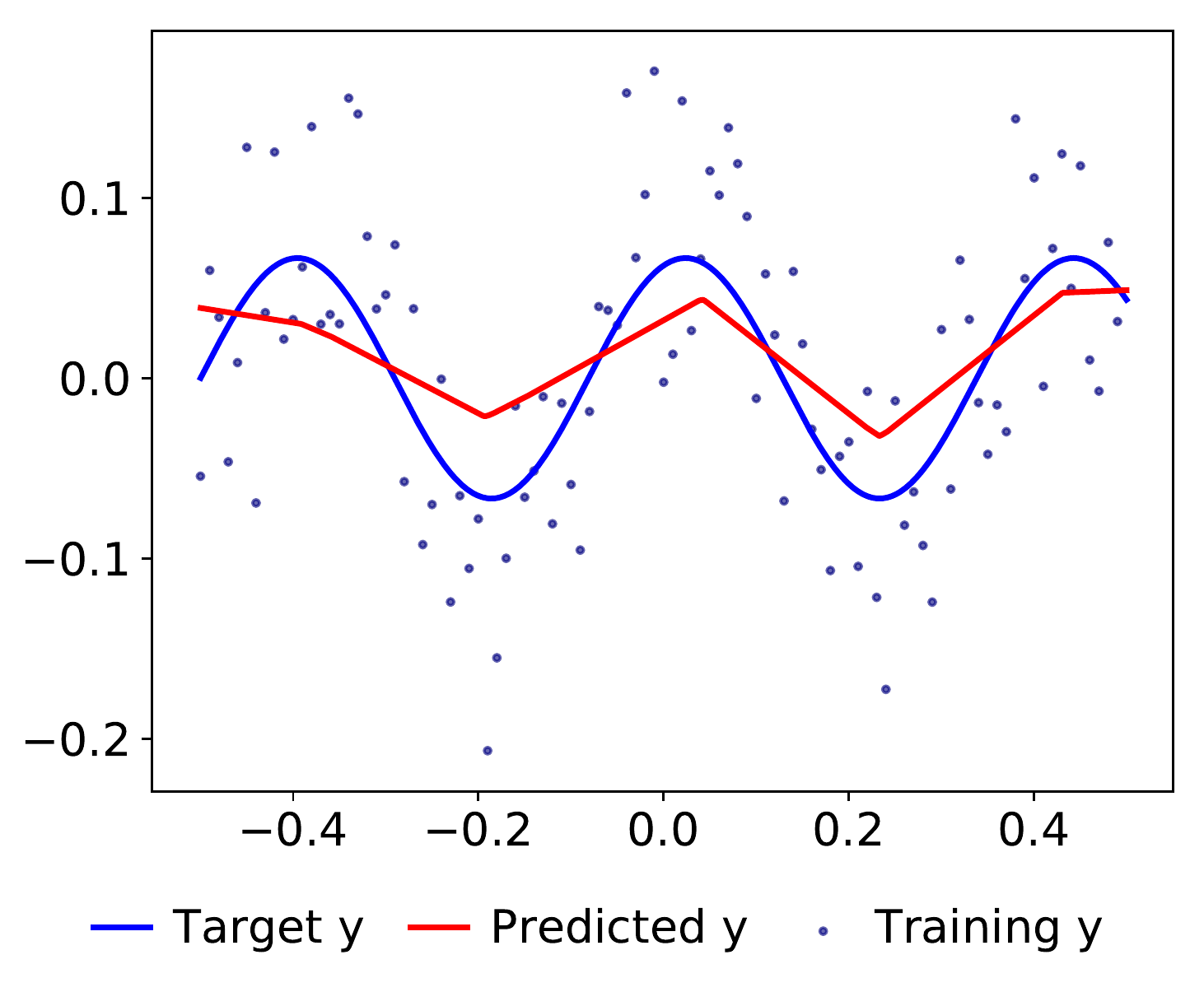}
    }
    \subfloat[bjorckGroupSort]
    {
        \includegraphics[width=0.32\linewidth]{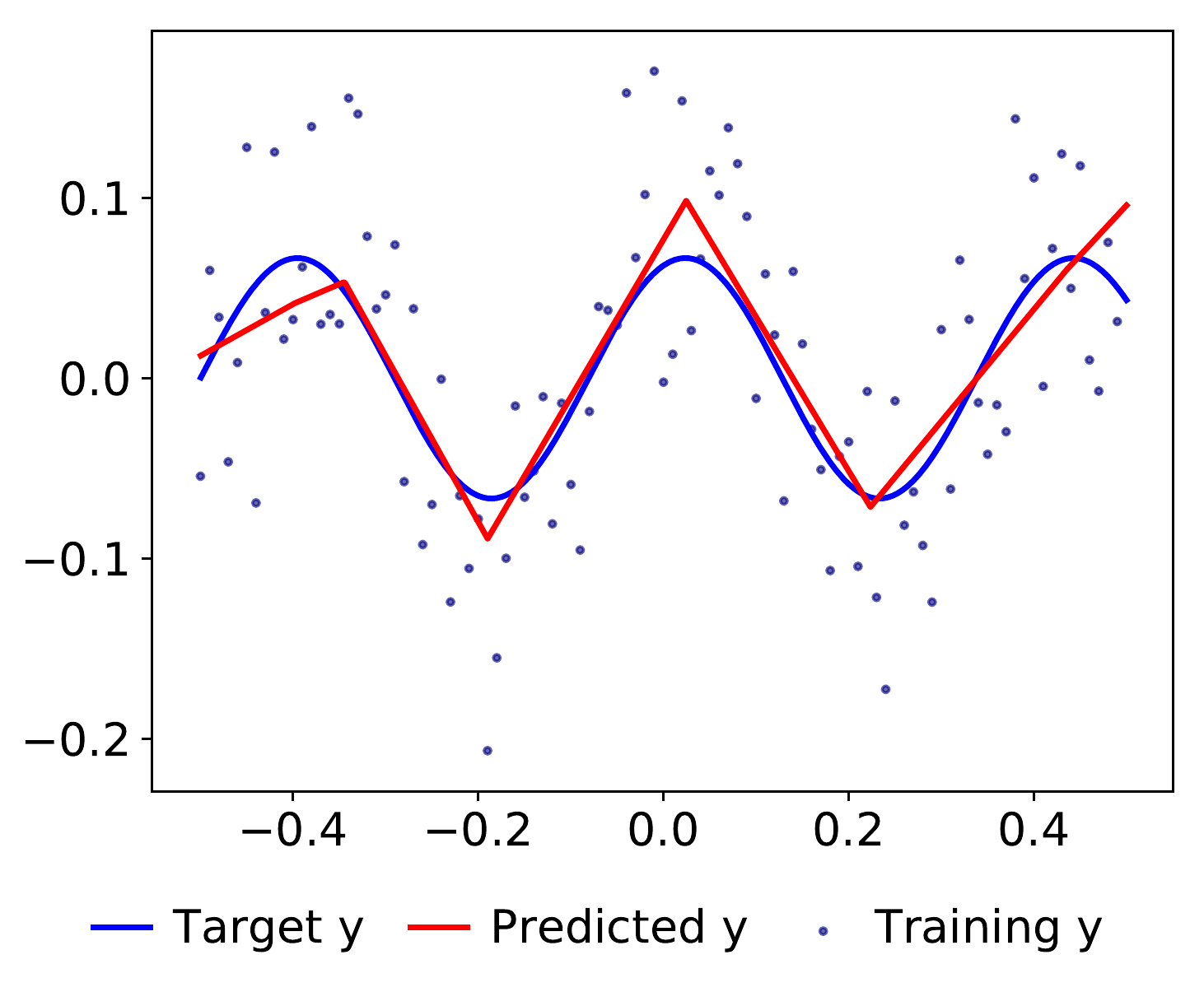}
    }
    \caption{Reconstructing the function $f(x)=(1/15)\sin(15x)$ in the model $Y=f(X)+\varepsilon$, with a dataset of size $n=100$.}
    \label{fig:approx_sinus_with_noise}
\end{figure}

\paragraph{The sinus function.} We provide in this subsection additional details for the learning of the sinus function $f(x)=(1/15)\sin(15x)$ defined on $[0,1]$ (see Section \ref{section:experiments}). Figure \ref{fig:approx_sinus_without_noise} is the case without noise while Figure \ref{fig:approx_sinus_with_noise} is the case with noise.

\subsection{Task 2: Calculating Wasserstein distances}
\begin{figure}[H]
    \centering
    \subfloat[$\mathscr{D}$ = ReLU network]
    {
        \includegraphics[width=0.31\linewidth]{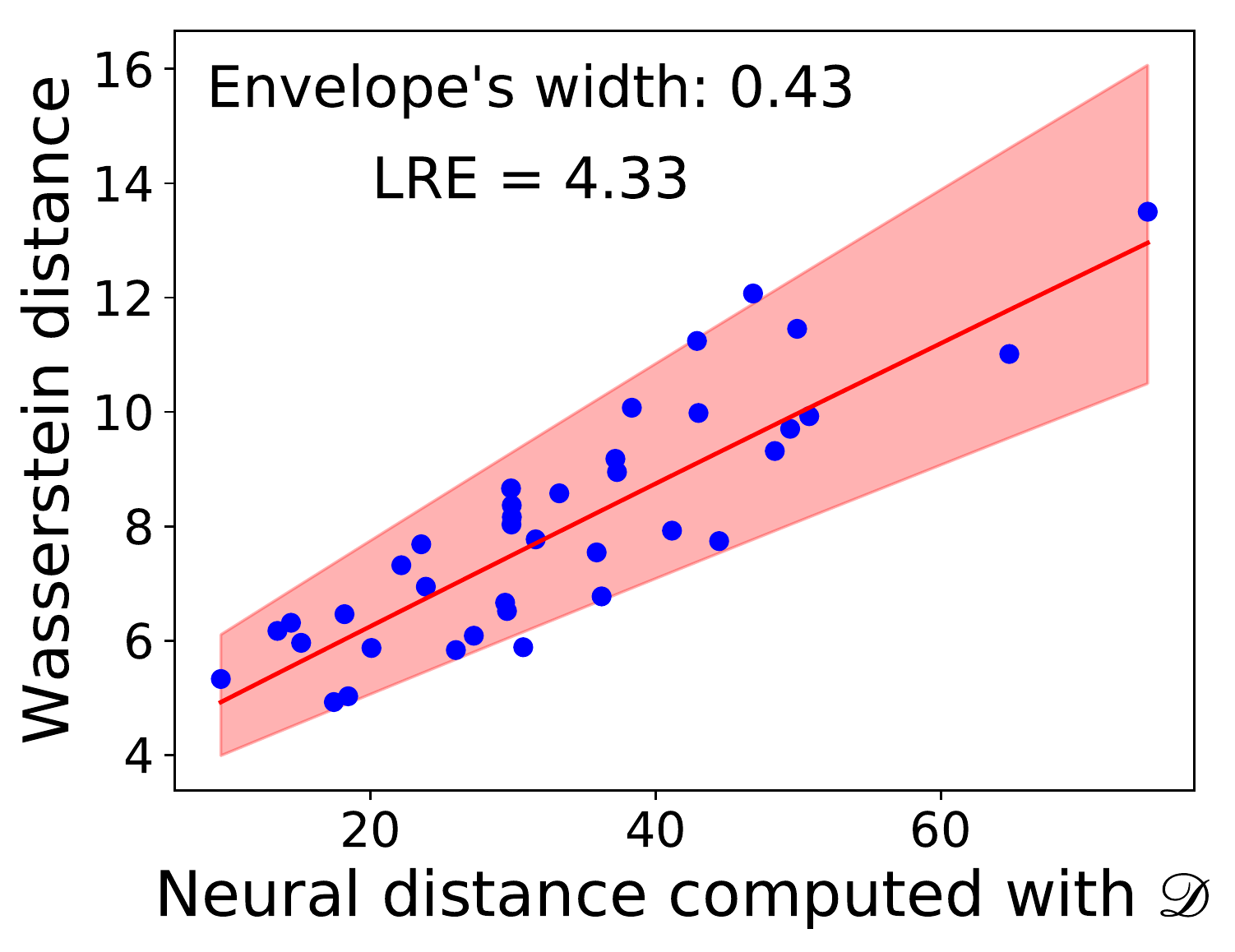}
    }
    \subfloat[$\mathscr{D}$ = bjorckReLU network]
    {
        \includegraphics[width=0.31\linewidth]{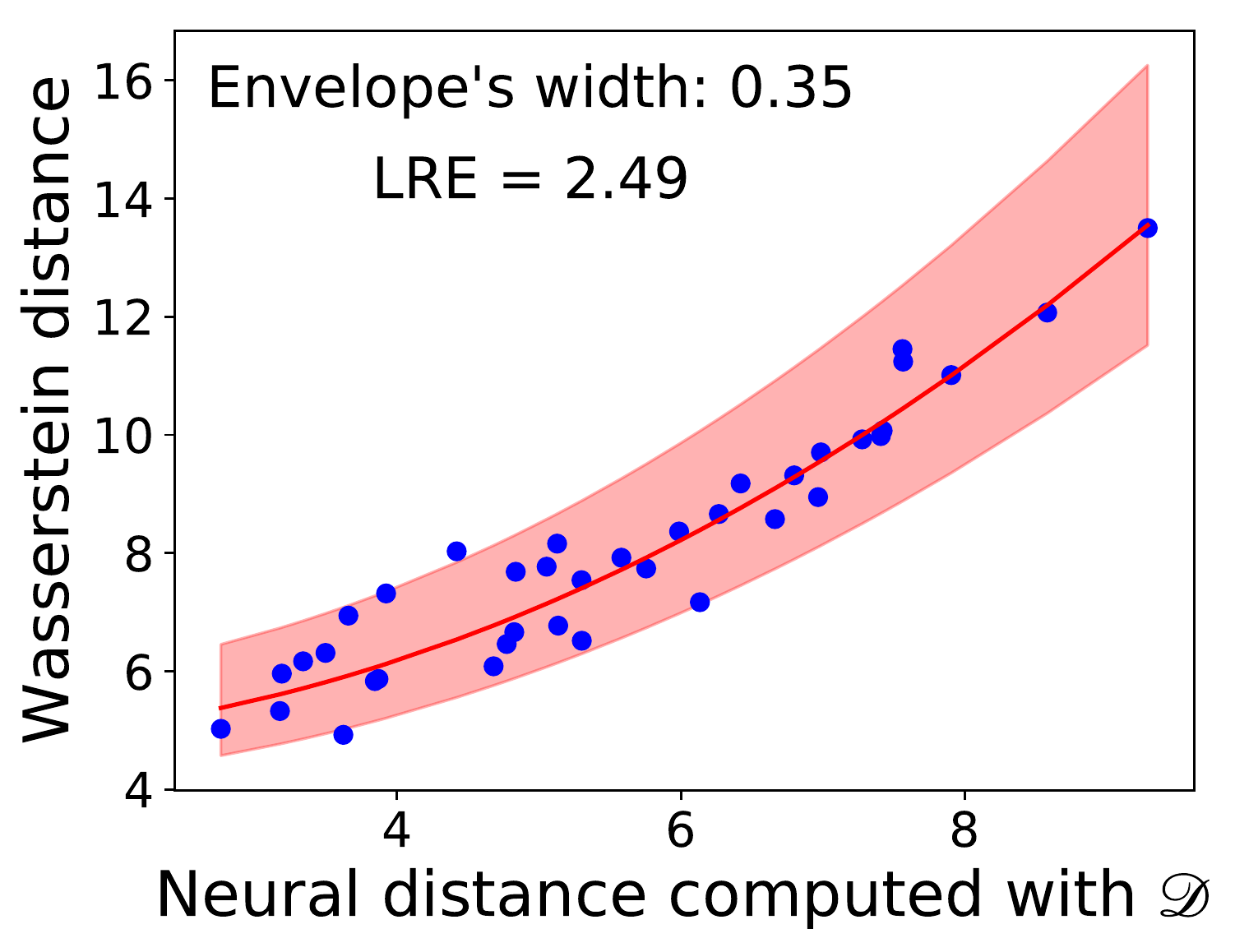}
    }
    \subfloat[$\mathscr{D}$ =  bjorckGroupSort network]
    {
        \includegraphics[width=0.31\linewidth]{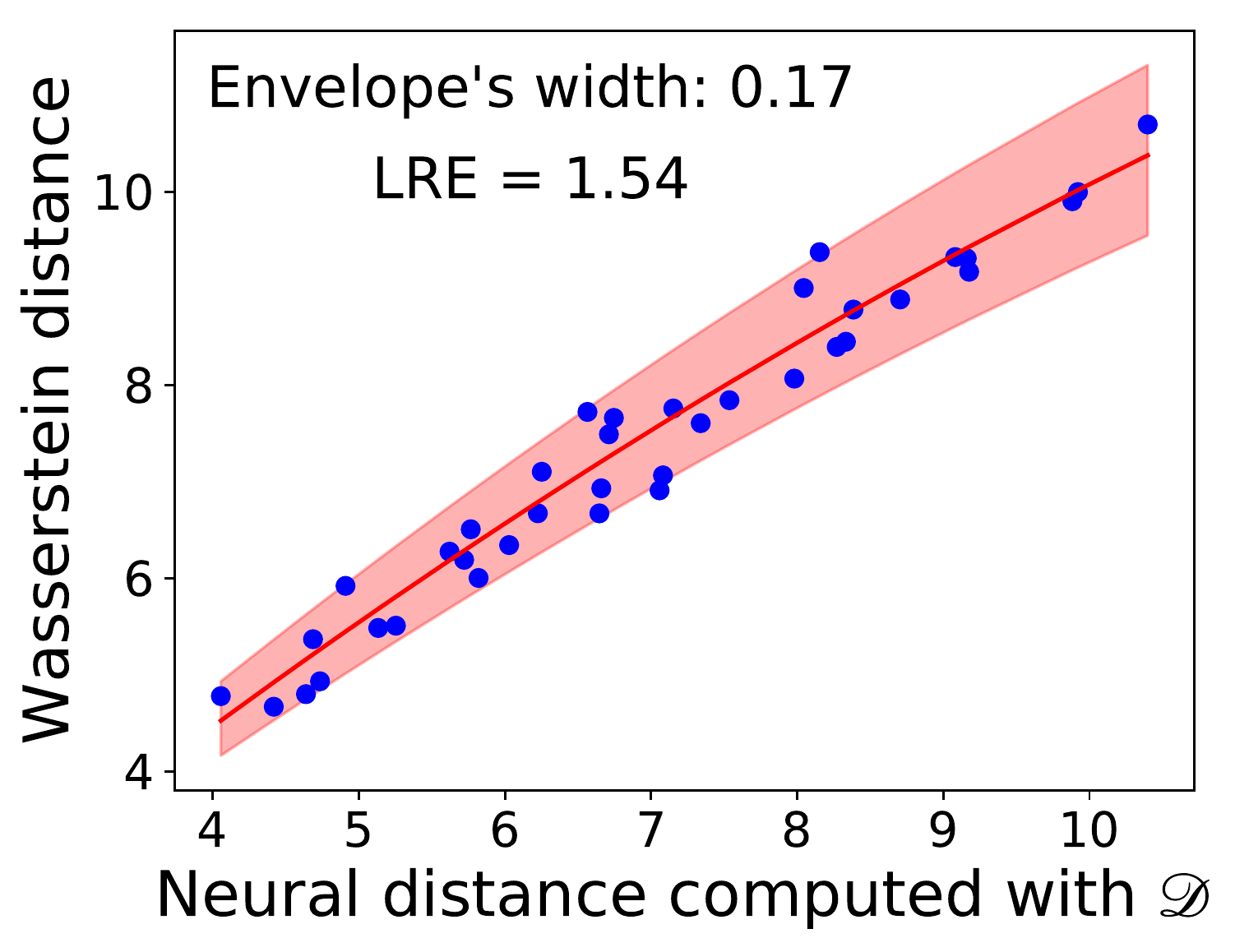}
    }
    \caption{Scatter plots of $40$ pairs of Wasserstein and neural distances, for $q=2$. The underlying distributions are bivariate Gaussian distributions with $4$ components. The red curve is the optimal parabolic fitting and LRE refers to the Least Relative Error. The red zone is the envelope obtained by stretching the optimal curve.}
    \label{fig:increasing_both_gen_and_disc_mnist_and_fashion_mnist}
\end{figure}

\section{Study of increasing group sizes for GroupSort networks}
\begin{figure}[H]
    \centering
    \subfloat[Grouping size = 2]
    {
        \includegraphics[width=0.31\linewidth]{images/PWL20_GS2.jpeg}
    }
    \subfloat[Grouping size = 5]
    {
        \includegraphics[width=0.31\linewidth]{images/PWL20_GS5.jpeg}
    }
    \subfloat[Grouping size = 10]
    {
        \includegraphics[width=0.31\linewidth]{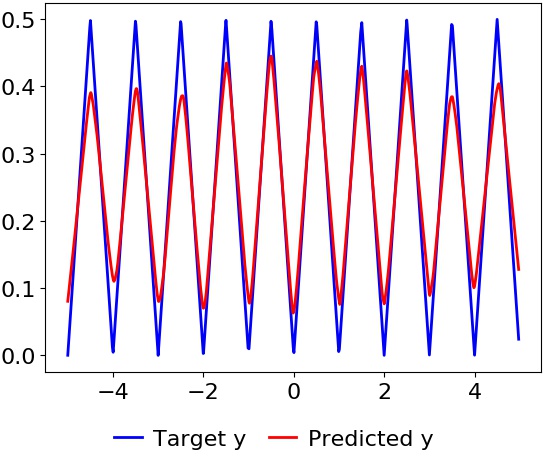}
    }
    \caption{Reconstruction of a $20$-piecewise linear function with varying grouping sizes ($k=2,5,10)$.}
    \label{fig:grouping_size1}
\end{figure}

\begin{figure}[H]
    \centering
    \subfloat[Grouping size = 2]
    {
        \includegraphics[width=0.31\linewidth]{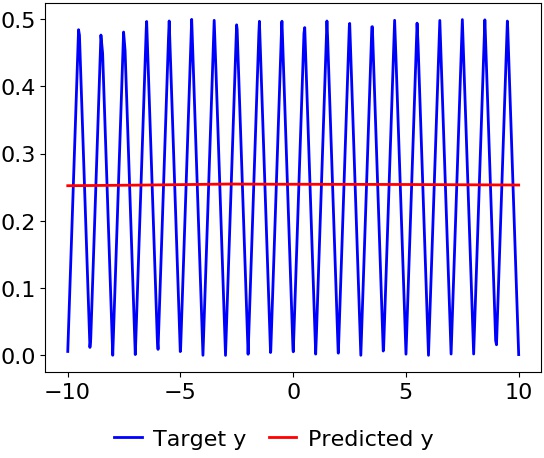}
    }
    \subfloat[Grouping size = 5]
    {
        \includegraphics[width=0.31\linewidth]{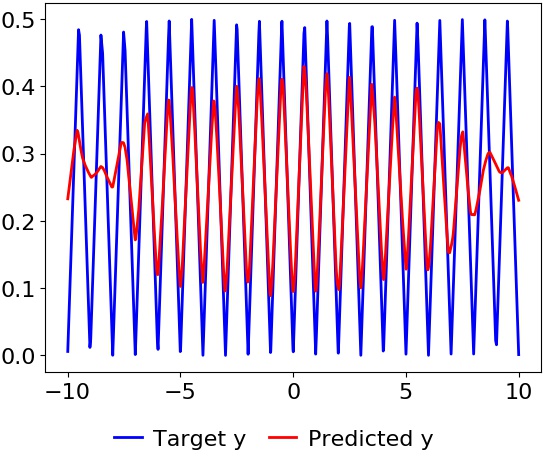}
    }
    \subfloat[Grouping size = 10]
    {
        \includegraphics[width=0.31\linewidth]{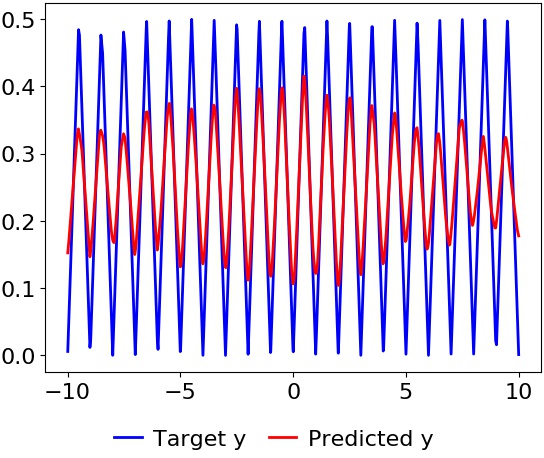}
    }
    \caption{Reconstruction of a $40$-piecewise linear function with varying grouping sizes ($k=2,5,10)$.}
    \label{fig:grouping_size2}
\end{figure}

\section{Shared architecture for both GroupSort and ReLU networks}
\begin{table}[H]
\centering
\resizebox{0.65\columnwidth}{!}{
\begin{tabular}{llllll}
    \toprule
    Operation & Feature Maps & Activation \\
    \midrule
    $D(x)$ \\
    Fully connected - $q$ layers & width $w$ & \{GroupSort, ReLU\} \\
    Width $w$ & \{50\} \\
    Depth $q$ & \{2, 4, 6, 8\} \\
    \midrule
    Batch size & 256\\
    Learning rate & 0.0025\\
    Optimizer & Adam: $\beta_1=0.5$ & $\beta_2=0.5$\\
    \bottomrule
\end{tabular}
}
\vspace{0.25cm}
\caption{Hyperparameters used for the training of all neural networks}
\label{appendix:tab:jsd_g_synthetic}
\end{table}

\end{document}